\documentclass[10pt,twocolumn,letterpaper]{article}
\usepackage[dvipsnames,svgnames]{xcolor}

\usepackage[pagenumbers]{cvpr} 

\usepackage{graphicx}
\usepackage{multirow}
\usepackage{tabularx}
\usepackage{booktabs}
\usepackage[pagebackref,breaklinks,colorlinks,citecolor=cyan]{hyperref}

\usepackage{color, colortbl}

\usepackage{arydshln}
\usepackage{pifont}

\title{VinTAGe: Joint Video and Text Conditioning for Holistic Audio Generation}

\author{%
    Saksham Singh Kushwaha, Yapeng Tian\\
    \textit{The University of Texas at Dallas}\\
}

\begin{document}
\maketitle
\begin{abstract}

Recent advances in audio generation have focused on text-to-audio (T2A) and video-to-audio (V2A) tasks. 
However, T2A or V2A methods cannot generate holistic sounds (onscreen and off-screen). This is because T2A cannot generate sounds aligning with onscreen objects, while V2A cannot generate semantically complete (offscreen sounds missing). In this work, we address the task of holistic audio generation: given a video and a text prompt, we aim to generate both onscreen and offscreen sounds that are temporally synchronized with the video and semantically aligned with text and video. Previous approaches for joint text and video-to-audio generation often suffer from modality bias, favoring one modality over the other. 
To overcome this limitation, we introduce VinTAGe, a flow-based transformer model that jointly considers text and video to guide audio generation. Our framework comprises two key components: a Visual-Text Encoder and a Joint VT-SiT model. To reduce modality bias and improve generation quality, we employ pretrained uni-modal text-to-audio and video-to-audio generation models for additional guidance. 
Due to the lack of appropriate benchmarks, we also introduce VinTAGe-Bench, a dataset of 636 video-text-audio pairs containing both onscreen and offscreen sounds. Our comprehensive experiments on VinTAGe-Bench demonstrate that joint text and visual interaction is necessary for holistic audio generation. Furthermore, VinTAGe achieves state-of-the-art results on the VGGSound benchmark. 
Our source code and pre-trained models will be released. Demo is available at: \url{https://www.youtube.com/watch?v=QmqWhUjPkJI}.

\end{abstract}    
\section{Introduction}
\label{sec:intro}

Post-production sound design is essential for adding realism and creating an immersive experience in silent videos, including films, music videos, animations, and games~\cite{chion2019audio,huiberts2010captivating}. While onscreen sounds provide direct information about key elements of the visual world, offscreen or invisible sounds—ubiquitous in scenes—depict environments beyond the immediate visual frame, enhancing the narrative while adding authenticity and emotional depth. From creating suspense in horror films, through unseen movements in the dark, to suggesting upcoming events, such as a distant roar of an approaching plane hinting at an airport setting, these offscreen sounds are typically added during post-production and are known as non-diegetic sounds. 

In practice, onscreen sounds are re-recorded using traditional Foley techniques~\cite{ament2014foley} or neural Foley methods~\cite{luo2023difffoley,zhang2024pia}, and then non-diegetic sound effects are created and mixed in. However, creating authentic audio mixes with both onscreen and offscreen sounds using traditional approaches~\cite{ament2014foley,izhaki2017mixing} requires specialized skills and is often labor-intensive. Despite recent progress in audio generation, producing a holistic audio experience that seamlessly blends visually aligned onscreen sounds with semantically relevant offscreen sounds remain a significant challenge.

\begin{figure}[tb]
    \centerline{\includegraphics[width=0.5\textwidth]{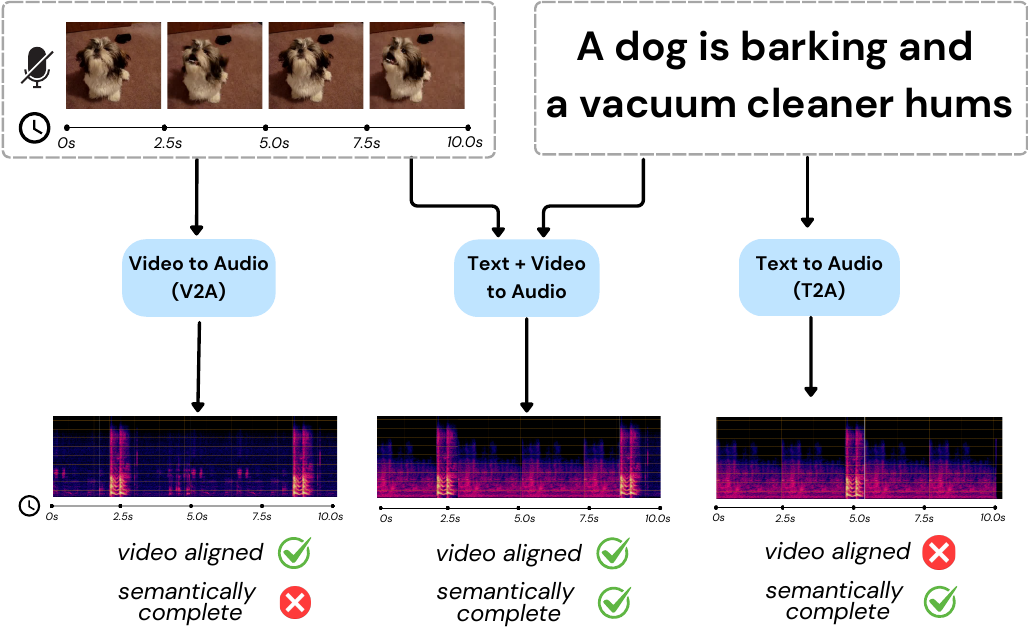}}
    \vspace{-2mm}
    \caption{Our VinTAGe model can generate visually aligned and text-corresponding sounds, including both onscreen and offscreen sound sources, providing a more holistic audio experience. However, V2A models can only generate video-aligned but semantically incomplete sounds, while T2A models, though generating semantically complete sounds, are not visually aligned.}
    \label{fig:teaser}
    \vspace{-5mm}
\end{figure}

Recent approaches to conditional audio generation broadly focus on text-to-audio (T2A)~\cite{audioldm2-2024taslp,ghosal2023tango,majumder2024tango,liu2023audioldm} and video-to-audio (V2A)~\cite{jeong2024read,luo2023difffoley,v2a-mapper,zhang2024pia} generation tasks. V2A methods aim to generate temporally and semantically synchronized audio with on-screen videos, while T2A models learn to generate high-quality audio that is semantically aligned with the provided text input. However, relying only on video or text as input is insufficient to generate a holistic audio experience. For example, as shown in Fig.~\ref{fig:teaser}, consider a household scene where a dog barks at approximately 2.5-second and 9-second in a 10-second video, while someone is vacuuming in the background. A V2A model may accurately generate the barking sounds at the appropriate times, but it would omit the ambient vacuum noise. Conversely, T2A model could produce a semantically complete soundscape, including the vacuum cleaner, but will fail to align the dog barking sounds temporally (\eg, placing them at the 5-second). Therefore, a model that considers both text and video modalities is essential for holistic audio generation.

In this work, we address the task of \emph{holistic audio generation} that is both visually aligned and text-driven. Given a video and a text prompt, our goal is to create high-quality audio that is temporally synchronised with video and semantically consistent with text, \ie, having onscreen and (if any) offscreen sounds. This task presents significant technical challenges, as it requires the model to jointly understand and integrate information from both visual and textual modalities to guide audio generation effectively. 
Existing approaches~\cite{zhang2024pia,jeong2024read,xie2024sonicvisionlm} that consider both text and video as conditions assume alignment between the input video and text. This assumption leads to modality bias and results in the generation of mostly onscreen sounds. However, in real-world scenarios, this assumption is often invalid, as offscreen sounds are prevalent and may not be directly related to the onscreen content—for example, an ambulance siren (an offscreen sound) might be heard in a video of a dog barking in a home setting.
Moreover, models like SonicVLM~\cite{xie2024sonicvisionlm} require explicit knowledge of onscreen and offscreen sounds and generate multiple audio tracks and hence necessitating manual mixing by a sound designer and impractical to scale. These limitations underscore the need of a method tailored to jointly interpret visual and textual inputs for holistic audio generation.
Additionally, a new benchmark is needed to effectively evaluate models on both onscreen and offscreen sound generation capabilities.

In this work, we propose VinTAGe, a novel flow-based transformer model for holistic audio generation guided by both video and text. Our architecture has two main components: the VT-Encoder and Joint VT-SiT model. The VT-Encoder efficiently encodes contextual embeddings from cross-interactions between video and text, incorporating motion and frame index information for temporal guidance. We extend the SiT model~\cite{ma2024sit} to condition jointly on the features generated by the VT-Encoder. To address modality bias, improve alignment, and generation quality, we introduce a teacher-student learning framework in which pretrained text-only and video-only audio generators serve as teachers, guiding our VT-SiT model through additional modality alignment losses.
Due to the lack of an appropriate benchmark for this task, we introduce VinTAGe-Bench, a new dataset to explore joint video-text to audio generation. VinTAGe-Bench consists of 636 (video, text, audio) pairs, consisting of 14 onscreen and 24 offscreen sound categories. The dataset includes 212 unique videos, each paired with one no-offscreen and two offscreen text captions, along with mixed audio tracks.

Through extensive objective and subjective evaluations, we found that jointly using both video and text features improves audio generation quality and alignment, highlighting the efficacy of the benchmark. Models that do not efficiently integrate text and video features tend to exhibit modality bias. Our carefully designed VT-Encoder and Joint VT-SiT model, enhanced with additional modality guidance losses, achieve superior performance. Furthermore, VinTAGe outperforms previous models on the VGGSound~\cite{Chen20} dataset.

Our contributions can be summarized as follows:
\begin{itemize}
    \item We introduce VinTAGe-Bench, a new benchmark to advance the under-explored task of holistic audio generation conditioned on both text and video inputs. 
    \item We propose VinTAGe, a novel flow-based transformer model specifically designed for this audio generation task, featuring a carefully structured VT-Encoder and Joint VT-SiT model, enhanced with teacher-student guidance.
    \item Through extensive evaluations, we demonstrate the effectiveness of our approach in generating high-quality audio that is both visually and contextually aligned.
    
\end{itemize}

\section{Related Works}

\textbf{{Text-to-Audio Generation.}} 
Initial works on text-to-audio generation, such as AudioGen~\cite{kreuk2022audiogen} and DiffSound~\cite{yang2023diffsound}, explored audio representation in discrete space given a text description, which is then decoded into waveforms. 
The advent of latent diffusion models~\cite{rombach2022high} inspired works like AudioLDM~\cite{liu2023audioldm}, AudioLDM2~\cite{audioldm2-2024taslp}, Tango~\cite{ghosal2023tango}, Tango2~\cite{majumder2024tango} and Make-An-Audio~\cite{huang2023make} for text-to-audio generation using latent diffusion techniques. AudioLDM employs an audio-text shared embedding space via CLAP~\cite{wu2023large}, utilizing audio data during training and text during inference. Tango~\cite{ghosal2023tango} improved audio generation with less data by using an instruction-tuned LLM FLAN-T5~\cite{chung2024scaling} text encoder for both training and inference. However, although these methods can generate high-quality audio, they alone cannot achieve temporal alignment with input visual content.

\noindent \textbf{{Video-to-Audio Generation.}} This field has shown promise in synthesizing sound for silent videos. A pioneering work by Zhou et al.~\cite{Zhou_2018_CVPR} proposed generating raw waveform samples from input video frames using SampleRNN~\cite{mehri2016samplernn}. SpecVQGAN~\cite{SpecVQGAN_Iashin_2021} synthesizes sound using RGB and motion features within a transformer-based autoregressive model. Im2Wav~\cite{sheffer2022i} uses image frame embeddings from the CLIP~\cite{radford2021learning} model to condition transformer models for autoregressive audio generation. Diff-Foley~\cite{luo2023difffoley} improves semantic and temporal synchronization by employing large-scale pretraining on audio-visual pairs and a latent diffusion model. However, these methods have mainly focused on temporal and semantic alignment with onscreen sounds.

Due to the prevalence of offscreen sounds, high-quality audio-visual pairs are scarce in real-world videos, as offscreen sounds are noisy for V2A generation models during training. Therefore, pretrained T2A generation models have been extended for V2A generation~\cite{xing24seeing,v2a-mapper,xie2024sonicvisionlm,jeong2024read,zhang2024pia,mo2024text}. ``Seeing and Hearing''~\cite{xing24seeing} uses an off-the-shelf pretrained ImageBind~\cite{girdhar2023imagebind} model to convert video frames into text, which is then fed into AudioLDM~\cite{liu2023audioldm}. Another approach, V2A-Mapper~\cite{v2a-mapper}, translates CLIP visual embeddings to the CLAP embedding space before using AudioLDM. To incorporate temporal guidance, SonicVLM~\cite{xie2024sonicvisionlm}, ReWaS~\cite{jeong2024read}, and FoleyCrater~\cite{zhang2024pia} train time-conditional estimators. However, these models generally assume that text and video are always aligned—which is not always true—and using a frozen text-to-audio model can lead to one modality overpowering the other.  In contrast, our proposed VinTAGe unifies visual and text modalities to simultaneously generate onscreen and offscreen sounds.

\begin{figure*}[t]
    \centering
    \includegraphics[width=0.99\textwidth]{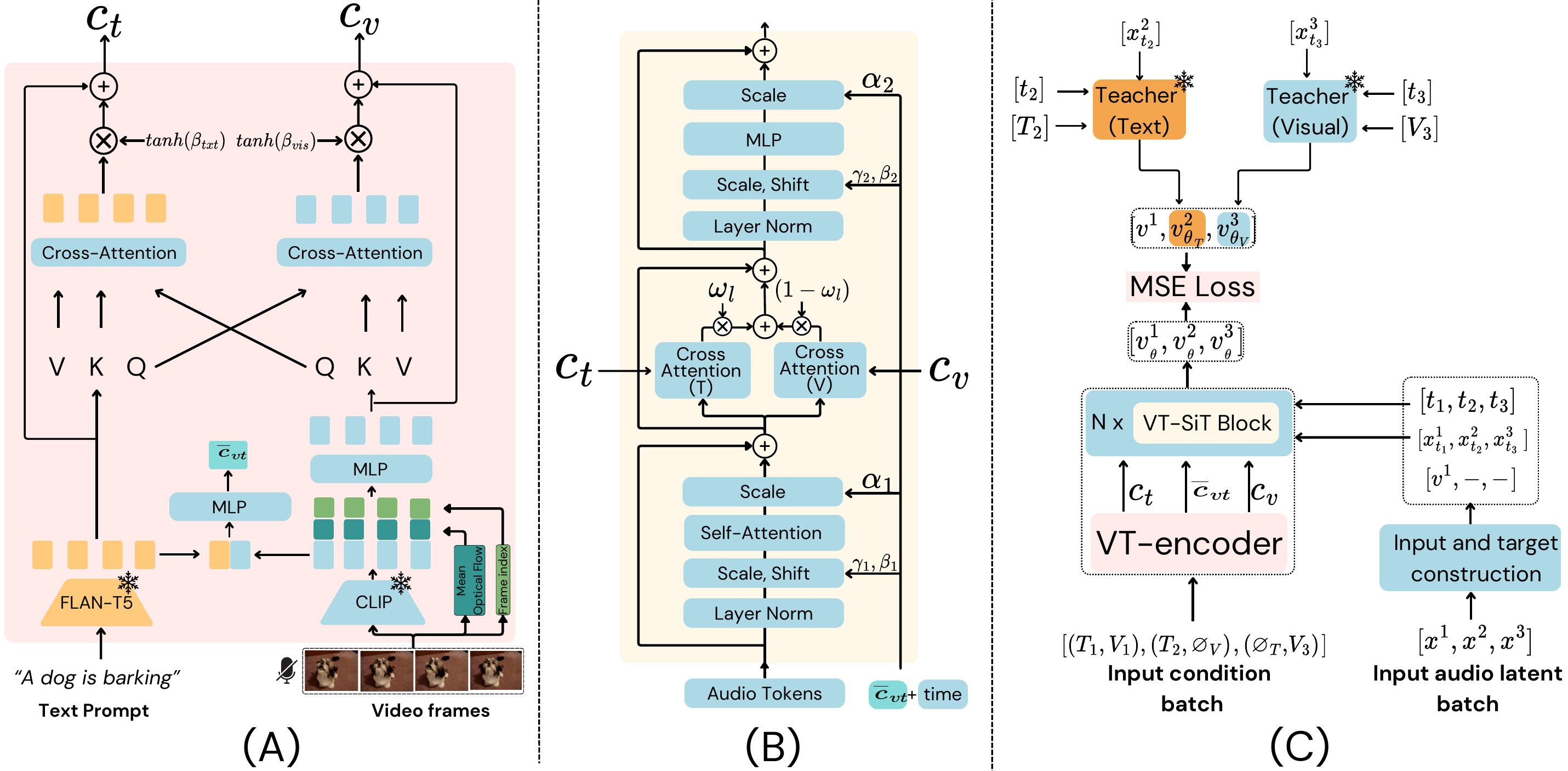}
    \caption{(A) VT-Encoder, (B) Joint VT-SiT block, (C) Overall training pipeline.}
    \vspace{-5mm}
    \label{fig:model}
\end{figure*}

\vspace{2mm}
\noindent
\textbf{Diffusion Transformers and Flow Matching.} 
Transformer-based diffusion models~\cite{ma2024sit,peebles2023scalablediffusionmodelstransformers,gao2024lumin-t2x} have gained increased interest due to their scalability and impressive results in image generation tasks. 
Flow matching models\cite{lipman2022flow}, which model the vector field of the transport probability path from noise to data, have demonstrated stable training and superior performance compared to score-based models. 
The Scalable Interpolant Transformers (SiT)~\cite{ma2024sit} employs a flow matching framework to improve the flexibility of connecting one data to another, further enhancing the performance of Diffusion Transformer (DiT) models~\cite{peebles2023scalablediffusionmodelstransformers}.
In the realm of audio generation, several models leveraging these advancements have been introduced, such as VoiceBox~\cite{le2023voiceboxtextguidedmultilingualuniversal}, AudioBox~\cite{vyas2023audioboxunifiedaudiogeneration}, FlashAudio~\cite{liu2024flashaudio}, LuminaNext~\cite{gao2024lumina-next}, and Diff-SAGe~\cite{kushwaha2024diff}. Building upon the advantages of flow matching in audio generation, we extend this approach to develop our joint visual-text to audio generation framework.

\section{Method}

\subsection{Preliminaries}
\label{prelim}

\noindent\textbf{Flow Matching.}
Flow matching~\cite{lipman2022flow,liu2022flow} linearly interpolates between noise and target data in a straight line. Specifically, given data $x \sim p(x)$ and Gaussian noise $\epsilon \sim \mathcal{N}(0, I)$, an interpolation-based forward process is defined as: \
\vspace{-1mm}
\begin{equation}
    x_t = \alpha_t x + \beta_t \epsilon,
\end{equation}

\vspace{-1mm}
where $\alpha_0 = \beta_1 = 1$, $\alpha_1 = \beta_0 = 0$, to satisfy this interpolation on $t \in [0,1]$ between $x_0 = x$ and $x_1 = \epsilon$. In our framework, we adopt the linear interpolation schedule between noise and original data, \emph{i.e.,} $ x_t = tx + (1-t)\epsilon.$

This formulation indicates a uniform transformation with constant velocity between data and noise. The corresponding time-dependent velocity field is given by 
\vspace{-1mm}
\begin{align}
    v_t(x_t) &= \dot\alpha_tx + \dot\beta_t\epsilon 
\end{align}

\vspace{-2mm}
where $\dot\alpha$ and $\dot\beta$ denote time derivative of $\alpha$ and $\beta$. 
This velocity can be approximated with a model $v_{\theta}(x_t, t)$ by minimizing the training objective:
\vspace{-1mm}
\begin{equation}
    \mathcal{L} = \mathbb{E}_{x,\epsilon,t}[\parallel v_{\theta}(x_t, t) - \dot\alpha_tx - \dot\beta_t\epsilon \parallel^2],
\label{eq:loss}
\end{equation}

\vspace{-1mm}
During inference, we integrate the probability flow ODE in reverse time, starting from a Gaussian noise sample \( x_1 \) at \( t = 1 \) and evolving it to \( t = 0 \) using an ODE solver such as Euler’s method. We iteratively update \( x_t \)  to obtain a final sample \( x_0 \), which aligns with the target data distribution.

\vspace{1mm}
\noindent\textbf{Audio VAE and Vocoder.}
Training generative models directly on waveform or mel-spectrograms ($m$) is computationally expensive. Hence previous approaches~\cite{ghosal2023tango,liu2023audioldm} have tackled this by training an autoencoder that compresses the spatial representation using Encoder $\mathcal{E}$. Latent generative models are trained on these compressed representations $x$ = $\mathcal{E}(m)$. During inference, the model generates latent $\hat{x}$, which are decoded back into mel-spectrograms $\hat{m} = \mathcal{D}(\hat{x})$ using the decoder $\mathcal{D}$ of the autoencoder. A vocoder is used to reconstruct waveform from synthesized sound spectrogram $\hat{m}$. Similar to previous approaches~\cite{ghosal2023tango}, we use Audio VAE and HiFi-GAN~\cite{kong2020hifi} vocoder from Liu et al.~\cite{liu2023audioldm}.

\subsection{Proposed Architecture}
\label{architecture}
Our architecture consists of three components: Visual and Text Encoder, Joint Visual-Text (VT) SiT  model, and Uni-modal teacher models. We will explain them one-by-one.

\noindent\textbf{Visual and Text (VT) Encoder.}
We employ a VT encoder module to generate embeddings from video frames and text inputs. Within this module as illustrated in Fig.~\ref{fig:model}A, the cross-modal embeddings interact and are associated to produce conditioning information for audio generation. 

\noindent
\textit{\underline{Text Encoder.}}
We adopt FLAN-T5~\cite{raffel2020exploring} as our text encoder to extract contextual language embeddings. A text prompt $T$ is encoded to $\tau\in\mathbb{R}^{L\times d_{txt}}$, where $L$ is the token count and $d_{txt}$ is the text embedding dimension.

\noindent\textit{\underline{Video Encoder.}} 
We utilize the visual encoder from CLIP~\cite{radford2021learning} to extract video representations. Given video frames $V \in \mathbb{R}^{N \times 3 \times H \times W}$, the CLIP visual encoder~\cite{radford2021learning} transforms them into embeddings $\mathcal{V}_{c} \in \mathbb{R}^{N \times d_{\text{clip}}}$, where each frame is represented by a $d_{\text{clip}}$-dimensional vector capturing its semantic content. To incorporate temporal information, we compute mean of optical flow~\cite{horn1981determining,yariv2024diverse} magnitude values sampled at the same rate as the video frames (N). We also include frame index numbers as explicit temporal position encodings. 
Both the optical flow values and frame indices are encoded using sinusoidal embeddings of  $d_{\text{opt}}$ and $d_{\text{idx}}$ respectively. We concatenate these embeddings with the CLIP embeddings $\mathcal{V}_{c}$, resulting in $\mathcal{V}_{\text{coi}} \in \mathbb{R}^{N \times (d_{\text{clip}} + d_{\text{opt}} + d_{\text{idx}})}$. Finally, we project the concatenated features using a linear layer for greater flexibility.

Previous approaches to adding temporal control have required extensive pretraining~\cite{luo2023difffoley} for temporally aligned features or involved training separate modules~\cite{zhang2024pia,jeong2024read,xie2024sonicvisionlm} for onset or energy detection from input video or expects user input~\cite{xie2024sonicvisionlm}. In contrast, our use of mean optical flow values provides a temporal energy map that is easier to get and directly guide audio generation.

\noindent
\textit{\underline{Visual-Text Cross-Attention.}}
To jointly generate audio from text and video, we utilize cross-attention between the text and visual embeddings to enable cross-modal association.
However, there may be scenarios where the two modalities are entirely unrelated or when we want to condition on them individually. To address this, we introduce a gated cross-attention mechanism with $\tanh$ gating~\cite{gao2024lumina} and skip connections, formulated as:

\vspace{-4mm}
\begin{align*}
c_{v} &= \mathcal{V}_{\text{coi}} + \tanh(\beta_{vis})\cdot\text{CA}(Q_{\text{vis}}, K_{\text{txt}}, V_{\text{txt}}) \\
c_{t} &= \mathcal{T} + \tanh(\beta_{txt})\cdot\text{CA}(Q_{\text{txt}}, K_{\text{vis}}, V_{\text{vis}})
\end{align*}
where \textit{$\text{CA}(Q, K, V)$} denotes the cross-attention operation with \textit{query} $Q$, \textit{key} $K$, \textit{and value} $V$.
In this mechanism, $\beta_{\text{txt}}$ and $\beta_{\text{vis}}$ are zero-initialized learnable parameters. They can help to incorporate cross-modal information gradually into the modality sequences and hence helps in stable training.
With the semantic visual and text embeddings, we can integrate them to generate a joint semantic text-visual condition vector: $\bar{c}_{vt} = \text{MLP}\left( \left[ \text{AvgPool}(\mathcal{V}_{c}); \ \text{AvgPool}(\mathcal{T}) \right] \right)$.

This joint pooled semantic embedding $\bar{c}_{vt}$ (along with time embedding) will be leveraged for global modulation using adaptive layer normalization (AdaLN)~\cite{perez2018film}. Additionally, contextual embeddings ($c_v$,$c_t$) will directly interact with the audio latent via cross-attention in VT-SiT.

\noindent\textbf{Joint VT-SiT.}
Our audio generator is built upon image generation model: SiT~\cite{ma2024sit}. As illustrated in Fig.~\ref{fig:model}B, input noisy audio latent is passed through $N$ blocks of the VT-SiT model to generate the predicted velocity $v_\theta$. 

We enhance the SiT blocks by adding text and visual cross-attention layers to leverage both modalities in guiding audio generation. In addition to the AdaLN-based global modulation~\cite{peebles2023scalablediffusionmodelstransformers}, this provides direct interaction and contextual guidance via attention~\cite{vaswani2017attention}. We introduce learnable weight parameters $\omega_l$ in each block $l$ to automatically learn the importance of each modality's guidance. The updated intermediate representation $\tilde{x}^{in}$ in block $l$ is computed as:
\vspace{-1mm}
\begin{align*}
\tilde{x}^{in} = x^{in} &+ \omega_l \cdot \text{CA}(Q_{x^{in}}, K_{c_t}, V_{c_t}) \\
&+ (1 - \omega_l) \cdot \text{CA}(Q_{x^{in}}, K_{c_v}, V_{c_v}),
\end{align*}
where $x^{in}$ is the intermediate noisy audio latent before cross-attention, and $\tilde{x}^{in}$ is after cross-attention with text and video embeddings. The learnable parameter $\omega_l$ allows the model to balance the contributions from the text and visual modalities dynamically at each layer.

\noindent\textbf{Visual and Text Teachers.} 
Generative models often inherit biases present in the training data~\cite{parihar2024balancing}, which can adversely affect their performance. We observe that previous works~\cite{zhang2024pia,jeong2024read} using text and video for audio generation suffer from \emph{modality bias}, tending to prioritize visual signals over textual input. To mitigate this issue, we employ pretrained teacher models that are individually trained for text-to-audio and video-to-audio generation tasks. These teacher models share architectures similar to our VT-SiT model, including cross-attention mechanisms and adaptive layer normalization. By utilizing these teacher models, we guide the joint visual-text audio generation process to effectively balance both modalities. Detailed architectural descriptions are provided in the appendix.

\subsection{Training and Inference}
\label{pipeline}
\begin{figure}[tb]
    \centerline{\includegraphics[width=0.38\textwidth]{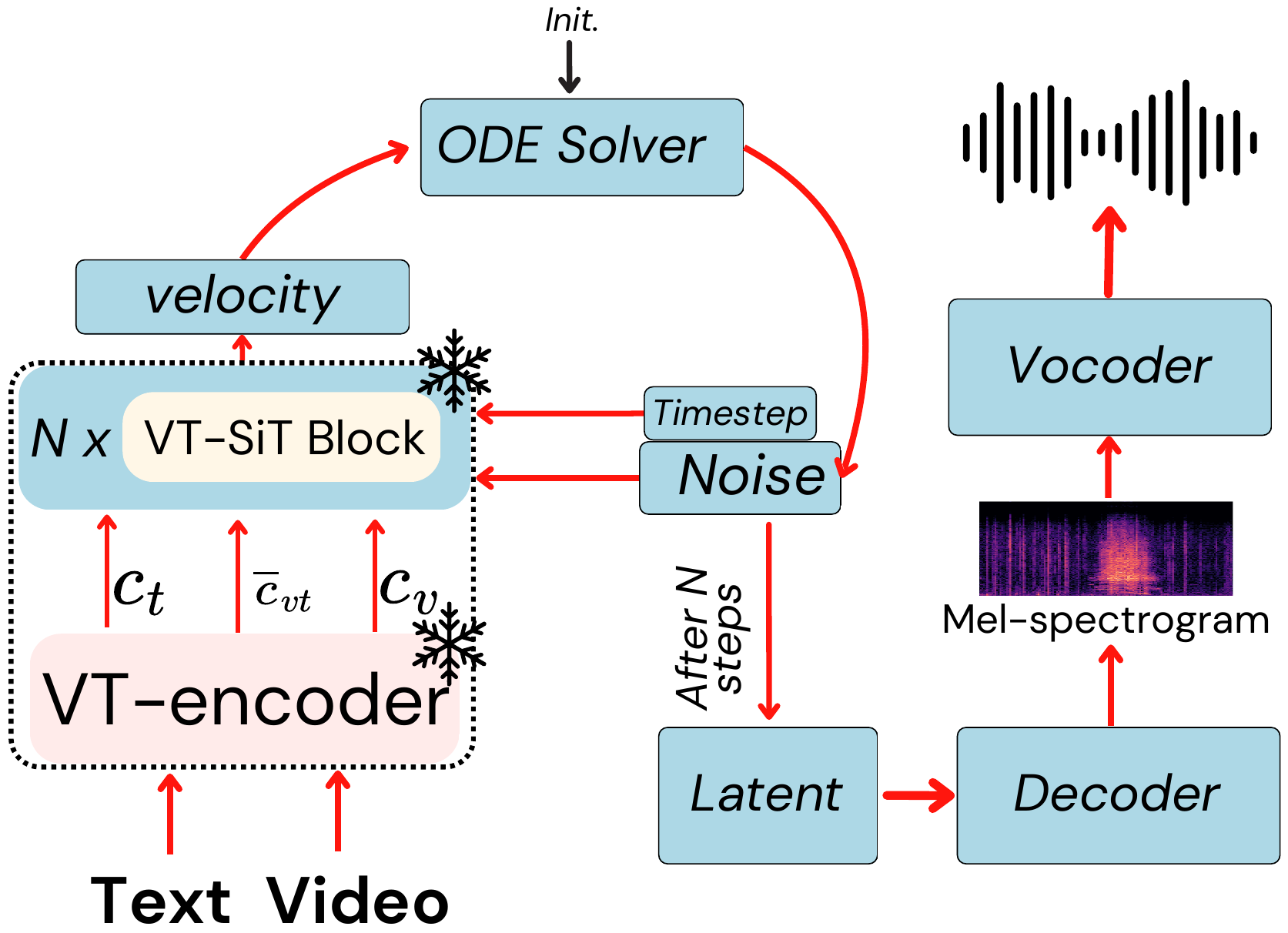}}
    \vspace{-2mm}
    \caption{Inference pipeline. }
    \label{fig:inference}
    \vspace{-5mm}
\end{figure}

In this subsection, we describe the training pipeline of VinTAGe, as illustrated in Fig.~\ref{fig:model}C, and explain the inference process using classifier-free guidance, shown in Fig.~\ref{fig:inference}.

Our task is to learn velocity field $v_\theta(x_t,t,c_{v},c_{t})$, given the noisy audio latent input $x_t$ interpolated at time t and given the visual and text contextual conditions: $c_{v}$ and $c_{t}$.

\noindent
\noindent\textbf{Creating Batches.} 
In order to avoid any modality bias during training, we use teacher models. In order to effectively use them each batch is carefully generated to maximize GPU memory and utilization. Each batch consists of a multiple of 3 inputs, where:

\begin{equation}
(T_i, V_i) =
\begin{cases}
(T_i, V_i) & \text{if } i \mod 3 = 0, \\
(T_i, \varnothing_V) & \text{if } i \mod 3 = 1, \\
(\varnothing_T, V_i) & \text{if } i \mod 3 = 2.
\end{cases}
\end{equation}

where \( T_i \) represents the text input, \( V_i \) represents the visual input, \( \varnothing_T \) indicates unconditioned text, and \( \varnothing_V \) indicates unconditioned visual input.

\begin{figure}[tb]
    \centerline{\includegraphics[width=0.4\textwidth]{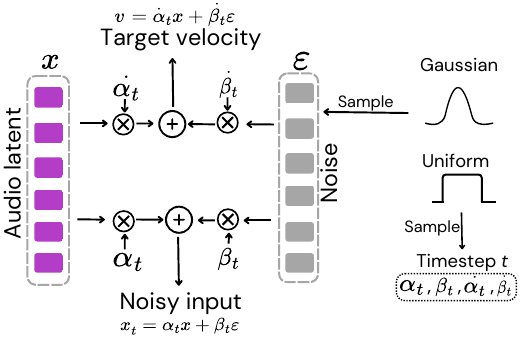}}
    \vspace{-2mm}
    \caption{Input and target construction. During training, an audio latent $x$ is interpolated at time t (sampled from uniform distribution), with random noise sampled from gaussion $\epsilon$ and generates noisy input and target velocity. }
    \label{fig:input_target_creation}
    \vspace{-3mm}
\end{figure}

\noindent\textbf{Input and Target Construction.} As shown in Fig.~\ref{fig:input_target_creation}, each audio latent $x_i$ is converted to noisy input and target velocity using Eqs. (1) and (2). In our case, we discard the target velocity for cases with $(\varnothing_T, V_i)$ and $(T_i,\varnothing_V)$ and replace them with the predicted velocity from teacher models. We observe that this not only helps in reducing modality bias but also improve generation quality.

\noindent\textbf{Teacher Guided Alignment.}
To mitigate modality bias and enhance the performance of our audio generation model, we introduce \emph{teacher-guided alignment losses}. This approach leverages pretrained teacher models that are individually trained on T2A and V2A generation tasks.

The alignment losses with the text-only teacher ($\mathcal{L}_{t}$) and the video-only teacher ($\mathcal{L}_{v}$) are defined as:
\vspace{-2mm}
\begin{align*}
\mathcal{L}_{t} &= \mathbb{E}_{x,\epsilon,t}\left[\left\| v_{\theta}(x_t, t, \varnothing_V, c_{t}) - v_{\theta_{\text{T}}}(x_t, t, c_{t}) \right\|^2\right], \\
\mathcal{L}_{v} &= \mathbb{E}_{x,\epsilon,t}\left[\left\| v_{\theta}(x_t, t, c_{v}, \varnothing_T) - v_{\theta_{\text{V}}}(x_t, t, c_{v}) \right\|^2\right],
\end{align*}

\vspace{-2mm}
where, $v_{\theta} $ is the velocity predicted by our joint model. $v_{\theta_{\text{T}}}$ and $v_{\theta_{\text{V}}}$ are the velocities predicted by the text- and visual-only teacher models, respectively.

The final objective combines the standard loss with these alignment losses:
\begin{align*}
\mathcal{L}_{\text{final}} = \mathbb{E}_{x,\epsilon,t}\left[\left\| v_{\theta}(x_t, t,c_{v}, c_{t}) - \dot{\alpha}_t x - \dot{\beta}_t \epsilon \right\|^2\right] \\
+ \lambda_v \mathcal{L}_{v} + \lambda_t \mathcal{L}_{t}.
\end{align*}

We find that incorporating the teacher-predicted velocities helps the model in two ways: regularization and improved quality. By comparing the joint model's outputs with those of the teacher models when only one modality is provided, we regularize the model to prevent dominance by either modality, thus mitigating the modality bias. Additionally, we observe that the model learns more effectively with teacher guidance than with the original velocity targets, resulting in higher-quality generated audio.

\noindent\textbf{Augmentation.}
To handle cases where text and video may not be fully aligned, we extend T2A augmentation methods~\cite{ghosal2023tango} by mixing audio, text, and video samples in various combinations, effectively increasing the diversity of our training data and enhancing the model's generation ability. More details are provided in the appendix.

\noindent\textbf{Inference using classifier-free guidance.}
Our model learns to predict the velocity conditioned on both text and visual inputs. To enhance the quality of the generated audio, we leverage \emph{classifier-free guidance} based on both conditions. During training, in addition to dropping text and visual conditions for teacher alignment, we also independently drop text and visual conditions with a 10\% probability to facilitate classifier-free guidance.
Following previous multi-conditional works~\cite{chen2024gentron,brooks2023instructpix2pix}, during inference we introduce two guidance scales, \( s_{\text{vis}} \) and \( s_{\text{txt}} \), to control the influence of each modality. The modified velocity prediction is computed as:
\begin{align*}
\tilde{v}_{\theta}(x_t, c_{v}, c_{t}) &= v_{\theta}(x_t, \varnothing_V, \varnothing_T) \\
&\quad + s_{\text{vis}} \cdot \left( v_{\theta}(x_t, c_{v}, \varnothing_T) - v_{\theta}(x_t, \varnothing_V, \varnothing_T) \right) \\
&\quad + s_{\text{txt}} \cdot \left( v_{\theta}(x_t, c_{v}, c_{t}) - v_{\theta}(x_t, c_{v}, \varnothing_T) \right),
\end{align*}
where $s_{\text{txt}}$ and $s_{\text{vis}}$ represent the weights of visual and text guidance.
As illustrated in Fig.~\ref{fig:inference}, we start the generation process from noise and, using the modified velocity function $\tilde{v}_{\theta}$, we perform \( N \) steps. The resulting latent representation is then passed through an Audio VAE Decoder and Vocoder to obtain the synthesized audio waveform. We empirically found $s_{\text{vis}}$=$s_{\text{txt}}$=2.5 to work well in our case and unless mentioned we use these values during inference.

\section{Experiments and Results}

\begin{table*}[t]
\centering
\caption{ Comparison on VinTAGe-Bench. Top-2 results are highlighted. }
\vspace{-2mm}
\resizebox{\textwidth}{!}
{\begin{tabular}{l|cc|ccc|cc>{\columncolor{Seashell}}c|cc>{\columncolor{Seashell}}c|ccc}
\toprule
\multirow{2}{*}{\textbf{Model}} & \multirow{2}{*}{\textbf{Txt}} & \multirow{2}{*}{\textbf{Vis}} & \multicolumn{3}{c|}{\textbf{Generation Quality}} & \multicolumn{3}{c|}{\textbf{Alignment}} & \multicolumn{3}{c|}{\textbf{Concept Accuracy}(\%)} & \multicolumn{3}{c}{\textbf{Subjective Metrics}} \\
\cmidrule(rl){4-6} \cmidrule(rl){7-9} \cmidrule(rl){10-12} \cmidrule(rl){13-15}
 & & & \textbf{FAD}$_\downarrow$ & \textbf{FID}$_\downarrow$ & \textbf{MKL}$_\downarrow$ & \textbf{AT}$_\uparrow$ & \textbf{AV}$_\uparrow$ & \textbf{Mean}$_\uparrow$ & \textbf{On-acc}$_\uparrow$ & \textbf{Off-acc}$_\uparrow$ & \textbf{Mean}$_\uparrow$ & \textbf{MOS-Q}$_\uparrow$ & \textbf{MOS-F}$_\uparrow$ & \textbf{MOS-T}$_\uparrow$ \\
\midrule
SpecVQGAN~\cite{SpecVQGAN_Iashin_2021} & \textcolor{OrangeRed}{\ding{55}} & \textcolor{ForestGreen}{\ding{51}} & 6.10 & 31.73 & 6.44 & 18.73 & 4.61 & 11.67 & 25.0 & 10.61 & 17.80 & 1.36 & 1.08 & 1.34 \\
Seeing-and-Hearing~\cite{xing24seeing} & \textcolor{OrangeRed}{\ding{55}} & \textcolor{ForestGreen}{\ding{51}} & 5.08 & 27.04 & 6.47 & 15.78 & 7.93 & 11.85 & 34.9 & 9.43 & 22.16 & - & - & - \\ 
Diff-Foley~\cite{luo2023difffoley} & \textcolor{OrangeRed}{\ding{55}} & \textcolor{ForestGreen}{\ding{51}} & 6.63 & 19.78 & 6.38 & 18.25 &  9.39 & 13.82 & 40.72 & 7.05 & 23.88 & 1.86 & 1.26 & 2.36 \\ 
\hdashline
Make-an-Audio~\cite{huang2023make} & \textcolor{ForestGreen}{\ding{51}} & \textcolor{OrangeRed}{\ding{55}}  & \textbf{4.05} & \textbf{19.01} & 5.19 & 18.54 & 7.42 & 12.98 & 52.04 & 35.14 & 47.83 & - & - & - \\
AudioLDM2~\cite{audioldm2-2024taslp} & \textcolor{ForestGreen}{\ding{51}} & \textcolor{OrangeRed}{\ding{55}}  & 5.40 & 20.75 & 5.52 &  22.02 & 6.65 & 14.33 & 54.4 & 26.65 & 40.52 & 2.42 & 2.52 & 2.10 \\
Tango2~\cite{majumder2024tango} & \textcolor{ForestGreen}{\ding{51}} & \textcolor{OrangeRed}{\ding{55}} & 5.85 & 36.01 & 4.94 & 23.84 & 7.19 & 15.51 & 62.57 & 48.58 & \textbf{55.57} & 2.86 & 3.34 & 2.56 \\

\hdashline 
Seeing-and-Hearing-VT & \textcolor{ForestGreen}{\ding{51}} & \textcolor{ForestGreen}{\ding{51}} & 4.89 & 22.44 & 5.00 & 17.75 & 8.99 & 13.37 & 43.23 & 17.68 & 30.45 & - & - & - \\
Tango2 + LLaVA~\cite{zhang2024llavanextvideo} & \textcolor{ForestGreen}{\ding{51}} & \textcolor{ForestGreen}{\ding{51}} & 4.12 & 29.1 & \textbf{4.59} & 24.14 & 7.35 & 15.75 & 57.86 & 40.33 & 49.09 & 2.88 & \textbf{2.90} & 2.52 \\
ReWaS~\cite{jeong2024read} &  \textcolor{ForestGreen}{\ding{51}} & \textcolor{ForestGreen}{\ding{51}} & 8.01 & 36.88 & 7.54 & 21.03 & 4.48 & 12.75 & 28.14 & 11.32 & 19.73 & - & - & - \\
FoleyCrafter~\cite{zhang2024pia} & \textcolor{ForestGreen}{\ding{51}} & \textcolor{ForestGreen}{\ding{51}} & 5.81 & 25.64 & 4.94 & 21.36 & 10.57 & \textbf{15.96} & 64.93 & 21.69 & 43.31 & \textbf{2.92} & 2.60 & \textbf{2.96} \\

\textbf{VinTAGe (Ours)} & \textcolor{ForestGreen}{\ding{51}} & \textcolor{ForestGreen}{\ding{51}} & \textbf{3.05} & \textbf{16.43} & \textbf{4.74} & 22.29 & 9.83 & \textbf{16.06} & 57.7 & 43.63 & \textbf{50.66} & \textbf{3.36} & \textbf{3.58} & \textbf{3.36} \\

\bottomrule
\end{tabular}
}
\vspace{-2mm}
\label{tab:main_table}
\end{table*}

\textbf{Datasets.} We adopt both VGGSound~\cite{Chen20} and our newly established VinTAGe-Bench in our experiments.

\noindent \underline{\textit{VGGSound.}} This dataset consists of $\sim$200K 10-second videos spanning 309 classes. Following prior V2A generation work~\cite{luo2023difffoley,SpecVQGAN_Iashin_2021}, we adopt the original train/test splits. Since the dataset includes only class labels, we enhance it with language descriptions by using LLM-generated captions from Auto-ACD~\cite{sun2024auto}. For any missing captions, we use a text prompt ``\textit{The sound of $\{$ class-label $\}$}.''

\begin{figure}[tb] \centering
    \includegraphics[width=0.46\textwidth,height=0.36\textwidth]{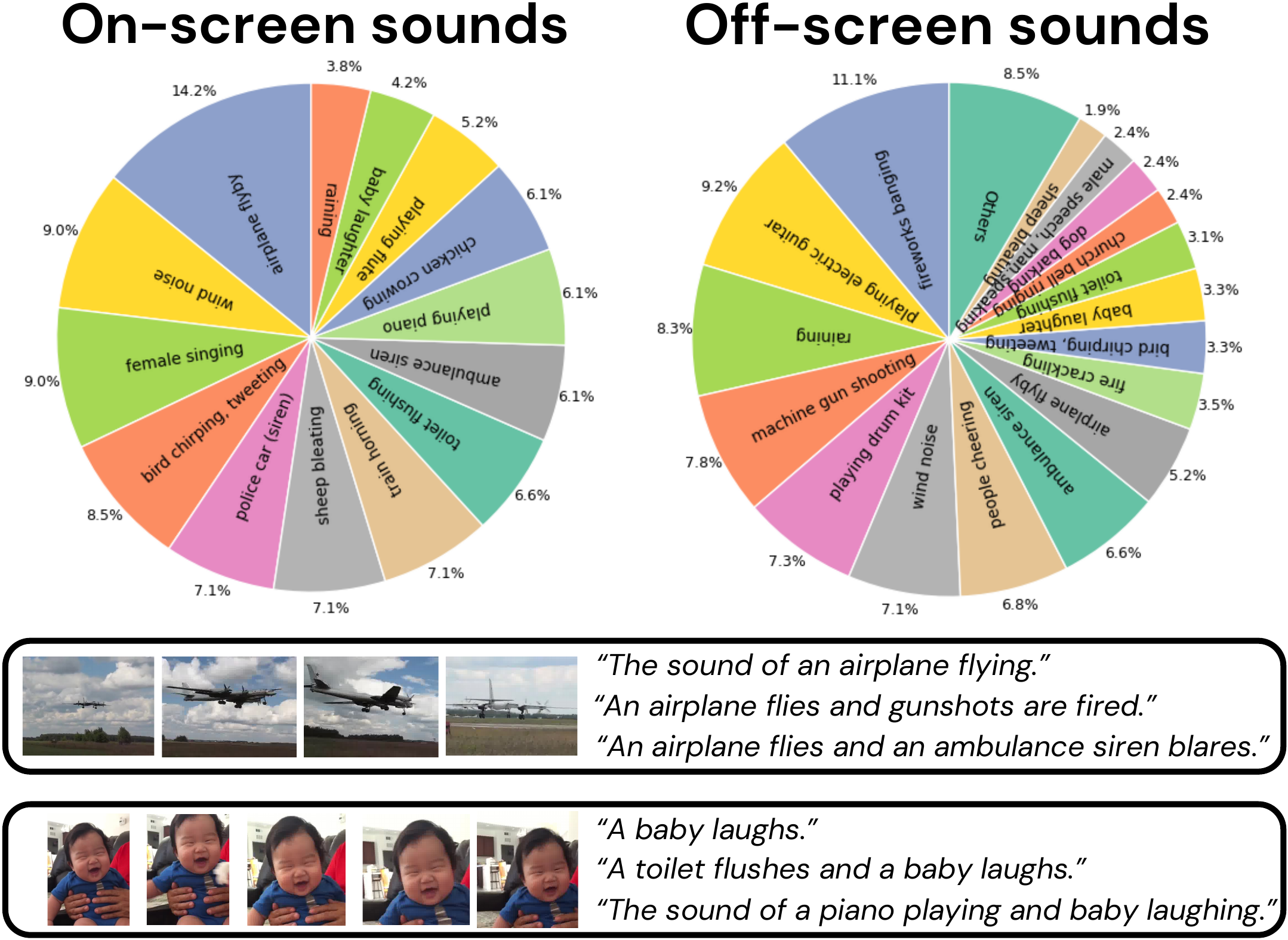}
    \vspace{-2mm}
    \caption{The statistics and examples of our VinTAGe-Bench. } \label{fig:bench_stats}
    \vspace{-5mm}
\end{figure}

\begin{figure*}[t]
    \centering
    \includegraphics[width=0.85\textwidth]{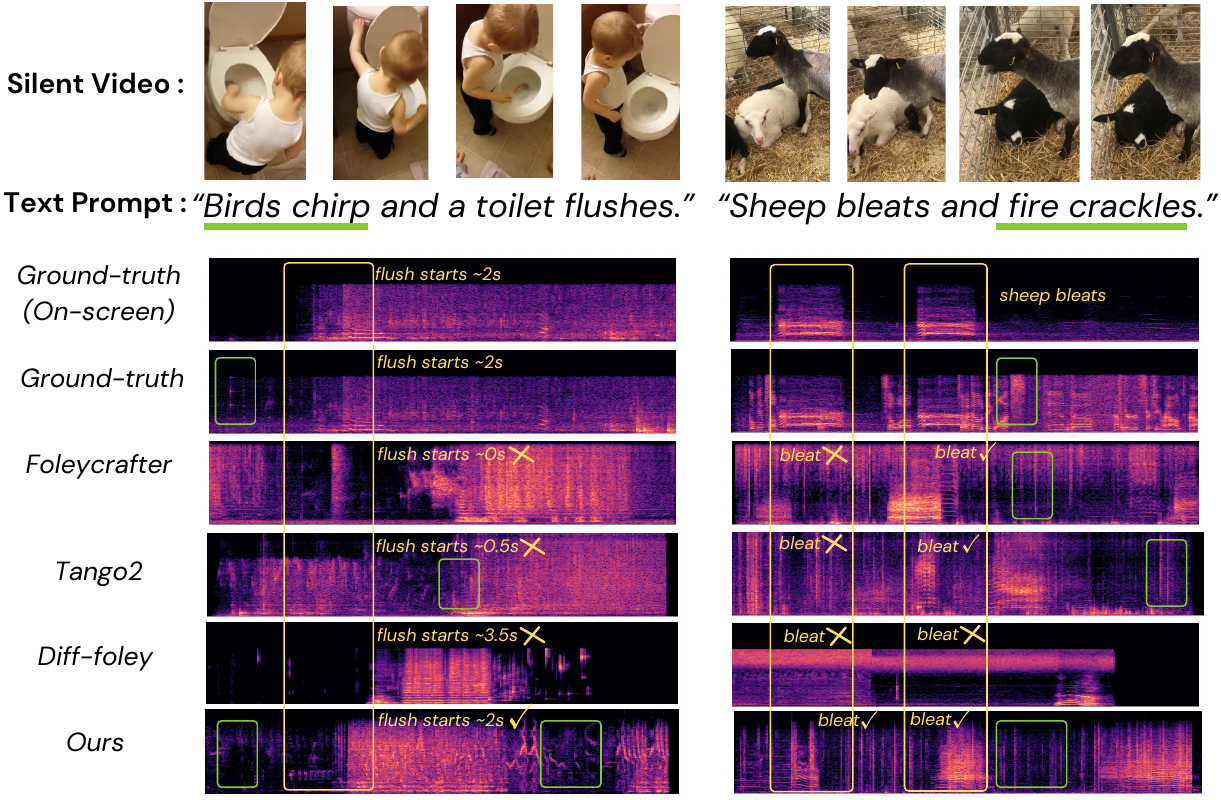}
    \caption{Examples of generated audio from VinTAGe-Bench. }
    \label{fig:qual_res}
    \vspace{-6mm}
\end{figure*}
\noindent \underline{\textit{VinTAGe-Bench.}} 
To the best of our knowledge, no suitable dataset exists for benchmarking audio generation conditioned on both text and video. We therefore created a dataset ensuring generated audio is both temporally synchronized with the video content and semantically consistent with accompanying text captions.

Based on the VGGSound test set, we selected 212 videos from 14 diverse classes, excluding those with limited object visibility, static images, and animations. For each class, we added offscreen sounds that could naturally co-occur or be less likely to co-occur with the onscreen sound. After mixing the audio, we filtered out scenarios where (1) Sounds with similar frequencies were mixed (\eg, vacuum cleaner and chainsaw), making them difficult to distinguish;
(2) Offscreen sounds were mixed with ``busy'' onscreen sounds (\eg, drums playing and dog barking), hindering recognition. Offscreen sounds were randomly selected from the VGGSound test set and mixed with the original audio, adjusting gains for balanced pressure levels~\cite{ghosal2023tango}.  Each video was paired with three scenarios: one with only onscreen sounds and two with added offscreen sounds.  Text captions were generated for each scenario, with the order of on-screen and off-screen sound descriptions randomly shuffled to prevent bias. This resulted in 636 (video, text,audio) pairs. The distribution of sounds is shown in Fig.~\ref{fig:bench_stats}.

\noindent
\textbf{Baselines.}
We compare our method with recent state-of-the-art V2A generation methods—SpecVQGAN~\cite{SpecVQGAN_Iashin_2021}, Seeing-and-Hearing~\cite{xing24seeing}, and Diff-Foley~\cite{luo2023difffoley}—and T2A methods like Tango2~\cite{majumder2024tango}, AudioLDM2~\cite{audioldm2-2024taslp}, and Make-an-audio~\cite{huang2023make}. We also include models that use both text and video inputs, such as FoleyCrafter~\cite{zhang2024pia} and ReWaS~\cite{jeong2024read}. To adapt single-condition models, we extend them by combining video and text captions. For Tango2, we generate video captions using LLaVa-Next~\cite{zhang2024llavanextvideo} and concatenate them with the original text captions, creating Tango2+LLaVa. Similarly, for Seeing-and-Hearing, we combine key-frame captions with original text captions to generate audio using AudioLDM, which we call Seeing-and-Hearing-VT.

\noindent \textbf{Evaluation Metrics.}
We employ both objective and subjective evaluations to measure audio generation performance.

\noindent\underline{\textit{Objective Evaluation.}}
We evaluate perceptual quality using Frechet Audio Distance (FAD)~\cite{kilgour2018fr} and Melception-based Frechet Distance (FID)~\cite{SpecVQGAN_Iashin_2021}, two key metrics for T2A and V2A generation tasks. Since FID and FAD capture distribution-level similarity, we also measure paired sample-level similarity with Mean KL Divergence (MKL)~\cite{SpecVQGAN_Iashin_2021}.
We evaluate audio-video (AV) and audio-text (AT) alignment on CLIP~\cite{radford2021learning} space by calculating cosine similarity times 100, following previous works~\cite{sheffer2022i,v2a-mapper,zhang2024pia}, using Wav2CLIP~\cite{wu2022wav2clip} as audio encoder and CLIP-text and image encoders for visual and text embedding. We compare the results on the mean of AV and AT scores for this joint task.
Additionally, we use a classifier to calculate if the generated audio faithfully captures both on-screen and off-screen concepts. Specifically, we separate on-screen and off-screen sounds based on text labels using Separate-Anything Model~\cite{liu2023separate} and then perform classification using ONE-PEACE~\cite{wang2023one}. Both are pretrained models with open-source and superior performance on VGGSound. We report mean results for this joint task.

\noindent\underline{\textit{Subjective Evaluation.}} To complement objective metrics, we conduct listening test to measure quality (MOS-Q), faithfulness (MOS-F) with both onscreen and offscreen sounds, and temporal alignment (MOS-T) of video with on-screen sounds. We ask 10 listeners to rate 35 samples on a discrete 5-point scale and report the mean opinion score. We also note that previous audio-visual temporal metrics~\cite{luo2023difffoley,yariv2024diverse} are not suitable in our case due to mixed audio containing offscreen sounds. Six baseline methods are compared.

\subsection{Comparison Results}
Our quantitative results on VinTAGe are shown in Tab.~\ref{tab:main_table}.

\noindent
\underline{\textit{T2A vs V2A.}} We observe that T2A models achieve superior audio generation quality compared to V2A models. One key reason is that there are more high-quality text-audio pairs and offscreen sounds are prevalent in videos.

\noindent
\underline{\textit{Joint Generation is Better.}} Combining visual and text features significantly enhances generation quality. For example, adding video captions to Tango2 improves all quality metrics, boosting FAD by $\sim$30\% by Tango2+LLaVA. This simple extension even surpasses FoleyCrafter and ReWaS, which also use text and video but less effectively. Our approach achieves the best FAD and FID scores, highlighting the superior quality of the synthesized audio.

\noindent
\underline{\textit{Audio-Text and Audio-Visual Alignment.}}  We observe that strong V2A models, such as Diff-Foley, generally achieve better AV alignment than T2A models but lack alignment with the text modality due to missing text information. Similarly, T2A models align well with text but fall short in visual alignment. Notably, Tango2+LLaVA improves upon Tango2 in mean alignment by adding video captions, while our approach achieves the best overall alignment. 

\noindent
\underline{\textit{Joint Generation Methods Suffer from Modality bias.}} For on- and off-screen accuracy, T2A models generally perform better, benefiting from large-scale, high-quality training data compared to V2A methods. Our mean concept accuracy is the second highest, slightly below Tango2 due to the limited quality of data, amount of data and to consider both text and video modalities. 
An interesting observation is that the two joint generation methods, ReWaS and FoleyCrafter, exhibit high on-screen accuracy but very low off-screen accuracy, indicating a strong visual bias. In contrast, our proposed approach effectively mitigates this issue.

\noindent
\underline{\textit{Subjective Study and Qualitative Results.}} Our approach outperforms other methods across all three subjective metrics, validating the quality, faithfulness, and temporal sync of the generated sounds. In Fig.~\ref{fig:qual_res}, we compare our approach with FoleyCrafter (VT2A), Tango2 (T2A), and Diff-foley (V2A). We see that our model can follow both onscreen sounds (\ie toilet flushing starting at $\sim$2 sec and sheep bleating twice) while other methods, even FoleyCrafter and Diff-foley, fail to follow.

\noindent
\underline{\textit{Comparison on VGGSound.}} We compare our approach on standard V2A generation benchmark: VGGSound~\cite{Chen20}, with commonly used V2A metrics as shown in Tab.~\ref{tab:vggsound}. We take the text captions from Auto-ACD~\cite{sun2024auto} and in order to have a fair comparison, we update previous text and video models (\eg, Foleycrafter and ReWaS) with these captions, which originally considered text labels. Our approach achieves the overall best performance. Subjective results also highlight good quality and temporal alignment. 

\begin{table}[tb]\centering
    \caption{Evaluations on VGGSound~\cite{Chen20} dataset.}
    \vspace{-3mm}
    \label{tab:vggsound}
    \resizebox{0.46\textwidth}{!}{
    \large
    \begin{tabular}{l|c|cccc|cc}
        \toprule
        \multirow{2}{*}{\textbf{Model}} & & \multicolumn{4}{c|}{\textbf{Objective metrics}} & \multicolumn{2}{c}{\textbf{Subjective metrics}} \\
        \cmidrule(rl){3-6} \cmidrule(rl){7-8} 
         & \textbf{FPS}$\downarrow$ & \textbf{FID}$\downarrow$ & \textbf{MKL}$\downarrow$ & \textbf{ISc}$\uparrow$ & \textbf{AV}$\uparrow$ & \textbf{MOS-Q}$\uparrow$ & \textbf{MOS-T}$\uparrow$  \\
        \hline
        SpecVQGAN (RGB+Flow)~\cite{SpecVQGAN_Iashin_2021} & 21.5 & 8.93 & 6.93 & 30.01 & 5.07 & - & - \\
        SpecVQGAN (ResNet50)~\cite{SpecVQGAN_Iashin_2021} & 21.5 & 9.70 & 7.03 & 30.80 & 5.87 & - & - \\
        Im2Wav~\cite{sheffer2022i}  & 30 & 11.44 & 5.20 & 39.30 & 7.82 & - & - \\
        Diff-foley~\cite{luo2023difffoley} & \textbf{4} & 9.87 & 6.43 & 62.37 & 9.17 & 3.03 & 3.46 \\
        ReWaS~\cite{jeong2024read} & 25 & 28.28 & 8.92 & 12.76 & 4.69 & - & - \\
        FoleyCrafter~\cite{zhang2024pia} & 15 & 9.17 & 4.48 & 62.49 & \textbf{10.13} & 3.80 & 3.17 \\
        \textbf{VinTAGe (Ours)}  & \textbf{4} & \textbf{6.65} & \textbf{4.12} & \textbf{63.95} & 9.68 & \textbf{3.86} & \textbf{3.88}\\
        
        \bottomrule
    \end{tabular}
    }
    \vspace{-3mm}
\end{table}

\begin{table}[tb]\centering
    \caption{Ablation on different model components.}
    \vspace{-3mm}
    \label{tab:ablate_main}
    \resizebox{0.43\textwidth}{!}{
    \begin{tabular}{l|cccc}
        \toprule
         & \textbf{FID}$\downarrow$ & \textbf{Mean-Align}$\uparrow$ & \textbf{Mean-Acc (\%)}$\uparrow$ & $\Delta{\textbf{On-Off}}(\%)\downarrow$  \\
        \hline
        VinTAGe & \textbf{16.43} & \textbf{16.06} & \textbf{50.66} & 14.07\\
        \hdashline
        \textit{w/o guidance loss} & 19.48 &  15.95 & 43.15 & 18.39 \\
        \textit{w/o CA VT-Encoder} & 18.80  & 16.04 & 41.66 & \textbf{13.52} \\
        \textit{w/o augmentation} & 19.22 & 15.65 & 38.55 & 20.99 \\
        \bottomrule
    \end{tabular}
    }
    \vspace{-5mm}
\end{table}
\section{Discussion and Analysis}

\begin{figure}[tb]
    \centerline{\includegraphics[width=0.4\textwidth, height=0.15\textwidth]{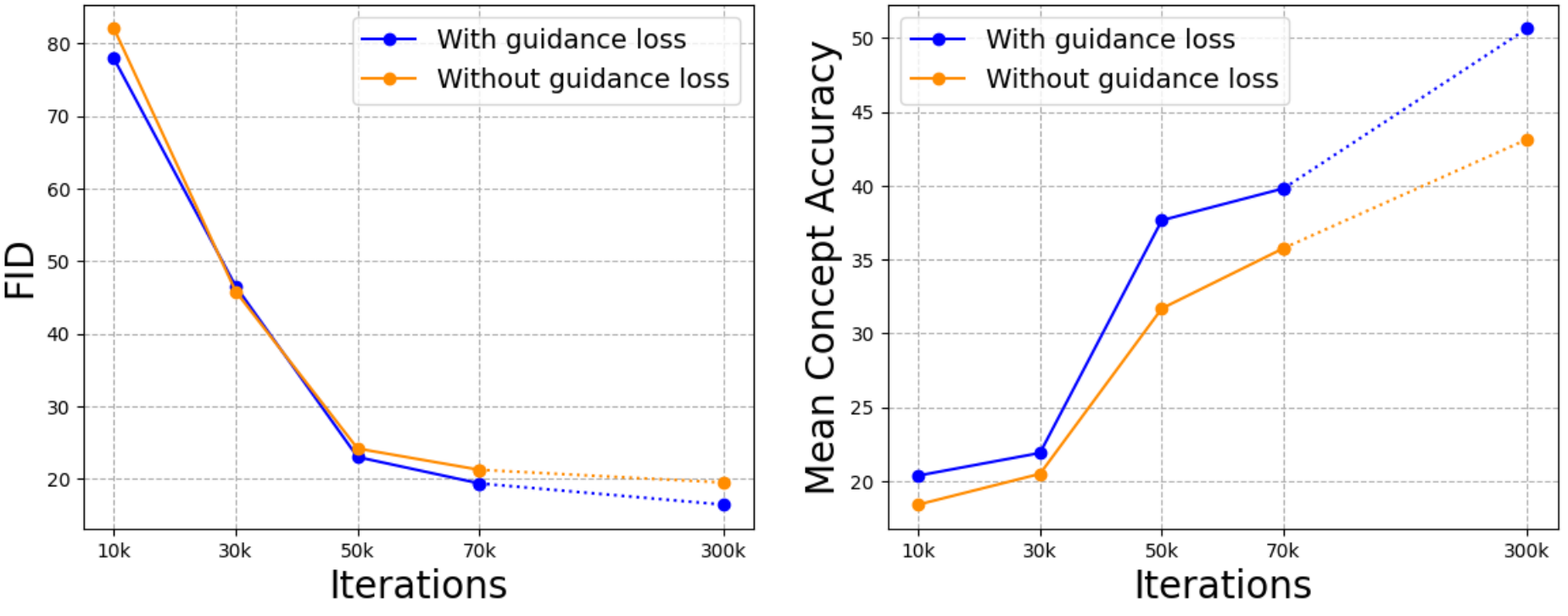}}
    \vspace{-2mm}
    \caption{Ablation on the alignment loss.}
    \label{fig:ablat_guidance}
    \vspace{-5mm}
\end{figure}

\noindent \textbf{Ablations on Model Components}. We do an ablation study to explore the effect of each component for our joint text-video to audio generation. From Tab.~\ref{tab:ablate_main}, we observe that each component positively affects the generation quality, alignment, and concept accuracy. Additionally, we find that augmentation is truly helpful in improving the concept accuracy. As shown in Fig.~\ref{fig:ablat_guidance}, guidance loss helps in learning better generation quality (FID) and concept accuracy (Mean-Acc), as it is easier to learn from teacher models.  We additionally see that the modality bias (On-Offscreen accuracy) is mitigated by guidance loss and augmentation. Furthermore, in Tab.~\ref{tab:ablate_vis}, we observe that mean-flow and frame-index features help to improve the generation quality.

\begin{table}[tb]\centering
    \caption{Ablation on V2A teacher model.}
    \vspace{-2mm}
    \label{tab:ablate_vis}
    \resizebox{0.38\textwidth}{!}{
    \begin{tabular}{l|cccc}
        \toprule
         & \textbf{FID}$\downarrow$ & \textbf{Mean-Align}$\uparrow$ & \textbf{Mean-Acc(\%)}$\uparrow$  \\
        \hline
        Teacher - V & \textbf{20.32} & \textbf{16.04} & \textbf{31.56} \\
        \hdashline
        \textit{w/o mean-flow} & 26.36 & 15.46 & 21.69  \\
        \textit{w/o frame-idx} & 25.94 & 15.30 & 21.02 \\
        \bottomrule
    \end{tabular}
    }
    \vspace{-3mm}
\end{table}

\begin{figure}[tb]
    \centerline{\includegraphics[width=0.43\textwidth]{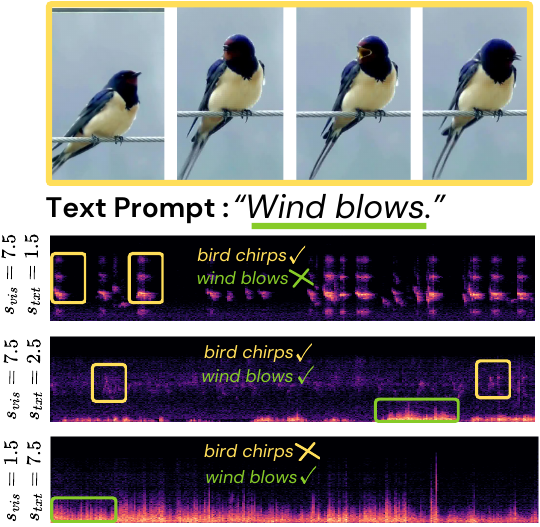}}
    \vspace{-2mm}
    \caption{Effect of classifier free guidance.}
    \label{fig:ablation_cfg}
    \vspace{-7mm}
\end{figure}

\noindent \textbf{Effect of $s_{vis}$ and $s_{txt}$}. 
We further analyse the effect of classifier-free guidance in Fig.~\ref{fig:ablation_cfg}. 
We observe that $s_{vis}$ controls the visual content and $s_{txt}$ controls the text content. For a simple scenario in which text and video are unrelated, higher $s_{vis}$ ($s_{vis}$=\textit{7.5},$s_{txt}$=\textit{1.5}) would result in only visual sounds and vice-versa. A good balance can result in a mixture of both the sounds ($s_{vis}$=7.5,$s_{txt}$=2.5).

\section{Conclusion}

In this paper, we tackle the task of holistic audio generation, aiming to produce audio that is both visually synchronized and aligned with text content. To address the lack of suitable datasets, we introduce VinTAGe-Bench—a curated dataset from the VGGSound, containing both onscreen and offscreen sounds. Additionally, we propose VinTAGe, a flow-based transformer model that jointly processes text and video inputs, leveraging V2A and T2A teacher models for guided learning. Extensive experiments and ablation studies validate the effectiveness of our approach.

{
    \small
    \bibliographystyle{ieeenat_fullname}
    \bibliography{main}
}

\appendix

\vfill

\vspace*{40\baselineskip}%
\section*{\centering \LARGE Appendix}

\section{Implementation details}

\begin{figure}[h!]
    \centerline{\includegraphics[width=0.4\textwidth]{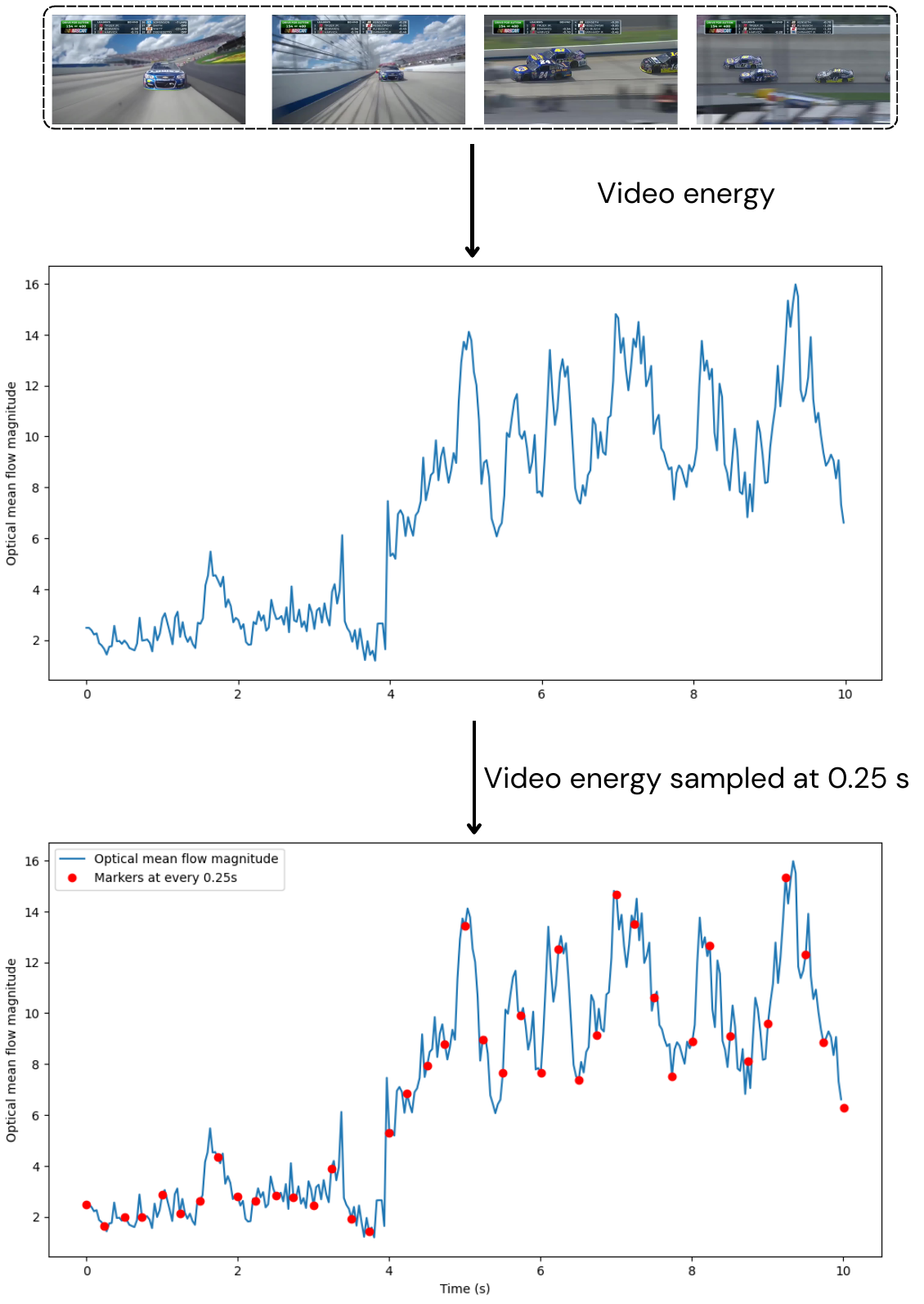}}
    \vspace{-2mm}
    \caption{To achieve temporal alignment with the video, we detect changes within the video by calculating the mean of the optical flow magnitude for each frame and sampling it to match the frame embedding rate.}
    \label{fig:energy}
    \vspace{-3mm}
\end{figure}

\subsection{Our method}

\noindent\textbf{Visual and Text encoder}: 
We employ a ViT-B/32-based~\cite{dosovitskiy2020image} pretrained CLIP model~\cite{radford2021learning} to extract frame features of dimension 512 at a sampling rate of 4 frames per second (fps). For each video, we obtain a CLIP embedding denoted as $\mathcal{V}_{c} \in \mathbb{R}^{40 \times 512}$. To achieve temporal alignment with the video, as illustrated in Fig.~\ref{fig:energy}, we detect changes within the video by computing the mean optical flow magnitude~\cite{horn1981determining,yariv2024diverse} for each frame. This information is then sampled to align with the extracted frame embeddings. 
Additionally, we include normalized frame indices $(0, 1, \ldots, 39)$ to represent frame positions. These video energy values and frame indices are encoded into sinusoidal embeddings of size 128 each and concatenated with the CLIP embedding, resulting in $\mathcal{V}_{\text{coi}} \in \mathbb{R}^{40 \times (512 + 128 + 128)}$. This concatenated embedding is subsequently passed through an MLP layer to produce a final embedding of size 1152.

We encode the text using the FLAN-T5~\cite{chung2024scaling} model with a text dimension of 1024, i.e., $\tau \in \mathbb{R}^{L \times 1024}$. 
In the cross-attention mechanism, we initialize $\beta_{\text{txt}} = \beta_{\text{vis}} = 0$ to allow for a gradual introduction of cross-modal information, thereby ensuring stable training. Additionally, we initialize $\omega_{l} = 0.5$ for each block $l$ of Joint VT-SiT, ensuring that both modalities contribute equally from the start of training.

We train our model on three L40 GPUs with an effective batch size of 9, utilizing the default optimization parameters as specified in SiT~\cite{ma2024sit,peebles2023scalablediffusionmodelstransformers}. Specifically, we build upon SiT-XL/2 model with patch size of $2 \times 2$, 28 blocks, and a hidden size of 1152. We train using the default SiT parameters, constant learning rate of $1 \times 10^{-4}$ with the Adam optimizer. All models are trained for 300K iterations. Also during training, we independently drop text and visual conditions with a probability of 10\%. We train the model for 5--6 days. 

\begin{figure}[tb]
    \centerline{\includegraphics[width=0.4\textwidth]{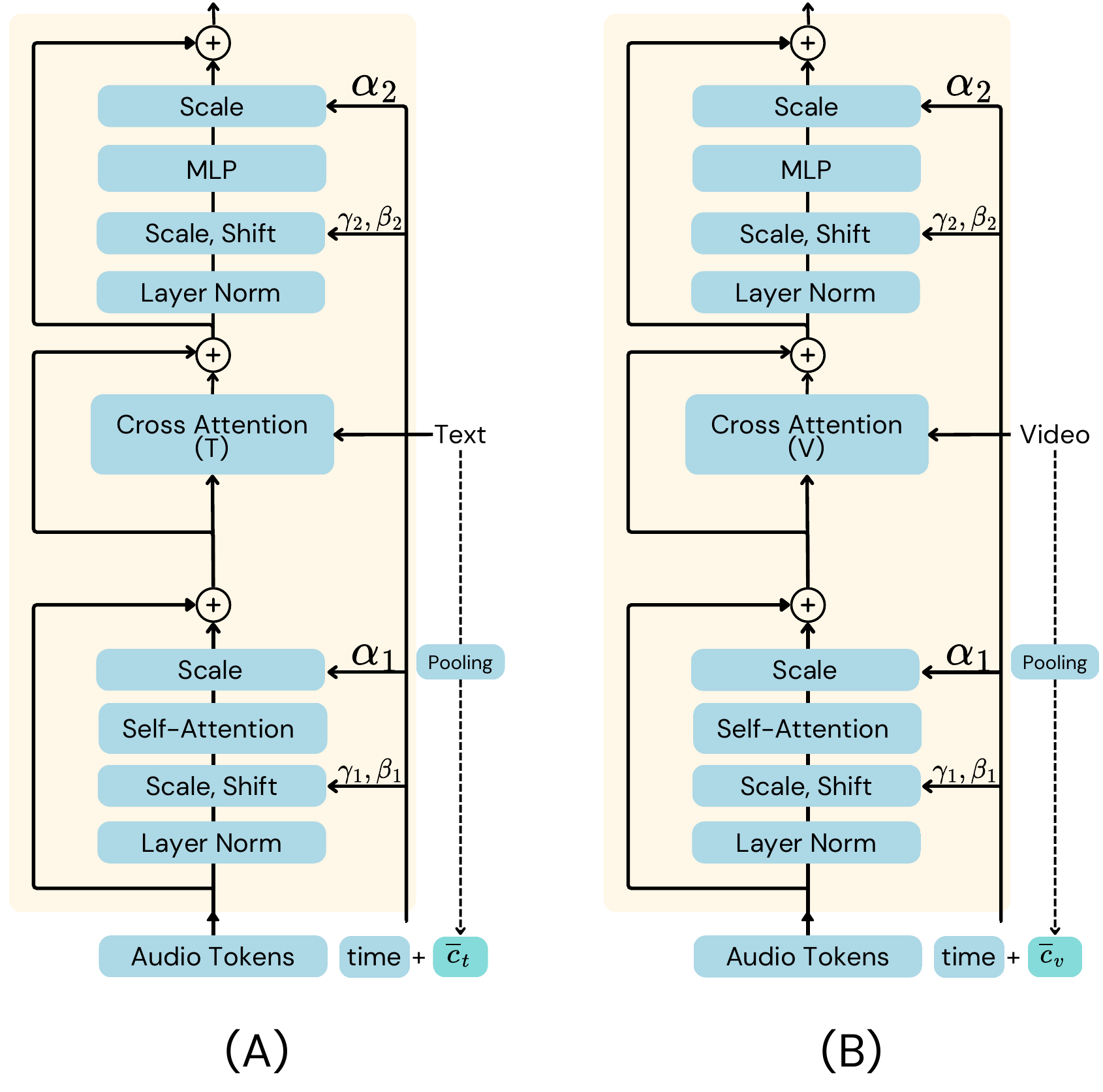}}
    \vspace{-2mm}
    \caption{Teacher models: (A) SiT block for text-only (B) SiT Block for visual-only to audio generation models}
    \label{fig:teacher}
    \vspace{-3mm}
\end{figure}

During the training of the Joint-VT SiT, we freeze the teacher models and set $\lambda_t = \lambda_v = 1$, utilizing the exponential moving average (EMA) weights of the teacher models. We present the architecture for the teacher models, which are trained for text-only and visual-only to audio generation. These teacher models are trained under the same settings as our VinTAGe model. The architecture of teacher model blocks are shown in ~\ref{fig:teacher}. 

During inference, we use the trained Joint VT-SiT model (trained on VGGSound training split) to perform evaluation on both the VGGSound test and VinTAGe-bench benchmarks. Unless otherwise specified, we set the scaling factors to $s_{\text{vis}} = s_{\text{txt}} = 2.5$. Similar to SiT~\cite{ma2024sit}, our sampling steps are 250.

\textbf{Audio processing.} 
We resample audio to 16\,kHz and pad shorter clips to 10 seconds. We extract spectrogram using an FFT size of 1024 and a hop size of 256, then convert it to a mel-spectrogram with dimensions $\mathbb{R}^{1 \times 1024 \times 64}$. This mel-spectrogram is fed into the Audio-VAE~\cite{liu2023audioldm} to obtain a latent representation of size $\mathbb{R}^{8 \times 256 \times 16}$. All our models are trained within this latent space. During inference, the generated latent is converted back to a mel-spectrogram (using Decoder of same Audio-VAE) and subsequently to audio waveform using HiFi-GAN~\cite{kong2020hifi}, following the approach of AudioLDM~\cite{liu2023audioldm}.

\textbf{Augmentations.} 
Standard datasets for audio generation typically consist of fully aligned audio, text, and video samples. However, during inference, text and video may be only partially aligned or even entirely unaligned. To address this discrepancy, we adopt data augmentation techniques from text-to-audio generation methods~\cite{ghosal2023tango,liu2023audioldm}, which enhance the diversity of our training data and enable our model to handle a broader range of multi-modal inputs. This strategy not only increases the effective dataset size but also improves the model's flexibility and overall generation quality.

Specifically, we effectively triple the number of data pairs through augmentation. Additionally, we mix audio samples based on their pressure levels~\cite{ghosal2023tango} to ensure that both audio sources remain audible. For example, given two data pairs $(v_1, t_1, a_1)$ and $(v_2, t_2, a_2)$ in a batch, we augment the dataset with the following combinations:
\begin{align*}
    (v_1, t_1 + t_2, \text{mix}(a_1,a_2)), \\
    (v_2, t_1 + t_2, \text{mix}(a_1,a_2)), \\
    (v_1, t_2, \text{mix}(a_1,a_2)), \\
    (v_2, t_1, \text{mix}(a_1,a_2)).
\end{align*}
Here, $(v, t, a)$ denotes the video, text caption, and audio, respectively. The expression $t_1 + t_2$ represents the concatenation of text captions \textit{"\{$t_1$\}"} and \textit{"\{$t_2$\}"}, while $\text{mix}(a_1,a_2)$ refers to the pressure-aware mixed audio.

To compute the mixed audio, we first calculate relative pressure level between $a_1$ and $a_2$. Let \(G_1\) and \(G_2\) denote the pressure levels of \(a_1\) and \(a_2\), respectively. We calculate the relative pressure level \(p\) as follows:
\[
    p = \left( 1 + 10^{\frac{G_1 - G_2}{20}} \right)^{-1}
\]
Following the methodology of~\cite{tokozume2017learning,ghosal2023tango}, the mixed audio is then computed by:
\[
    \text{mix}(a_1,a_2) = \frac{p \cdot a_1 + (1 - p) \cdot a_2}{\sqrt{p^2 + (1 - p)^2}}.
\]

\subsection{Baselines}
For all visual-to-audio and text-to-audio generation baselines, we following to their default parameters for audio generation. VGGSound~\cite{Chen20} and VinTAGe-Bench consist of 10-second videos. Since ReWaS~\cite{jeong2024read} generates audio for 5-second durations, we generate audio for 5secs chuncks and combine them to match the 10-second video length.
For Tango2+LLaVa, we generate video captions using LLava-Next~\cite{zhang2024llavanextvideo} by providing the prompt: \textit{"Answer the following question directly in one short sentence. What is the main content in the video?"}. The generated caption is then combined with the text prompt as \textit{"\{video caption\} and \{text prompt\}"} and passed to the Tango2~\cite{majumder2024tango} model for audio generation.

\section{Additional analysis}

\subsection{Input conditions.} We compare our model with teacher models in Tab.~\ref{tab:condition} to highlight the benefits of joint text-video guidance. Our model can also operate with video-only or text-only input by setting the respective weights to zero during classifier-free guidance. For fair comparison, we maintain a guidance scale (cfg) of 5 for individual conditioning and teacher models. Our VinTAGe model demonstrates efficiency by saving approximately 0.6 billion parameters, which is the difference between the sum of the teacher models' parameters and those of our model.

\begin{table}[tb]\centering
    \caption{Conditioning on both modality and Model Size comparison}
    \label{tab:condition}
    \resizebox{0.48\textwidth}{!}{
    \begin{tabular}{lc|cc|ccc}
        \toprule
         \textbf{Model} & \textbf{\#Param} & \textbf{Txt} & \textbf{Vis} & \textbf{FID}$\downarrow$ & \textbf{Mean-Align}$\uparrow$ & \textbf{Mean-Acc}$\uparrow$  \\
        \hline
        VinTAGe (2.5,2.5) & 1.3B & \textcolor{ForestGreen}{\ding{51}} & \textcolor{ForestGreen}{\ding{51}} & \textbf{16.43} & \textbf{16.06} & 50.66 \\ 
        \hline 
        Teacher - T (5.0) & 1.1B & \textcolor{ForestGreen}{\ding{51}} & \textcolor{OrangeRed}{\ding{55}} & 24.66 & 15.07 & 53.65 \\
        VinTAGe (0.0,5.0) & 1.3B & \textcolor{ForestGreen}{\ding{51}} & \textcolor{OrangeRed}{\ding{55}} & 21.18 & 14.97 & \textbf{54.32} \\
        \hline
        Teacher - V (5.0) & 0.8B & \textcolor{OrangeRed}{\ding{55}} & \textcolor{ForestGreen}{\ding{51}} & 20.32 & 16.04 & 31.56 \\
        VinTAGe (5.0,0.0) & 1.3B & \textcolor{OrangeRed}{\ding{55}} & \textcolor{ForestGreen}{\ding{51}} & 18.53 & 15.24 & 22.44 \\
        \bottomrule
    \end{tabular}
    }
\end{table}

\subsection{Text prompts.}
\begin{figure}[tb]
    \centerline{\includegraphics[width=0.49\textwidth, height=0.8\textwidth]{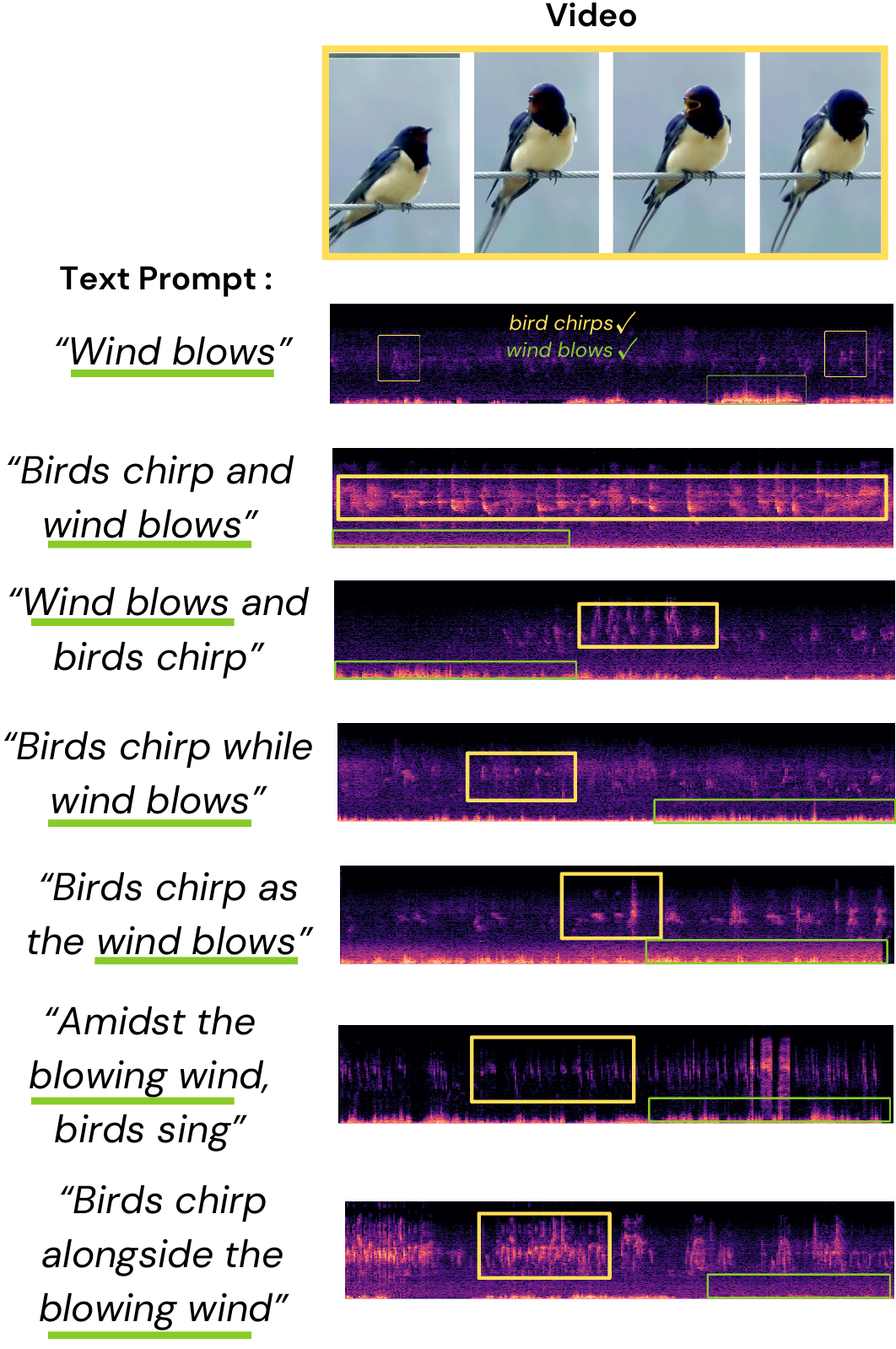}}
    \vspace{-2mm}
    \caption{Generation examples with different text permutation. }
    \label{fig:spec_abltation}
\end{figure}

Our model can handle several different scenarios of text prompt. As shown in Fig.~\ref{fig:spec_abltation}, we conduct qualitative tests on different combinations of text prompts. First, we evaluate the case where the text and video are completely unaligned. Additionally, we use ChatGPT to generate variations where "bird chirping" and "wind blowing" occur simultaneously. Our observations indicate that the model effectively manages multiple scenarios, demonstrating its robustness and flexibility. 

\subsection{Additional analysis on $s_{vis}$ and $s_{txt}$.}
\begin{figure}[h]
    \centerline{\includegraphics[width=0.48\textwidth]{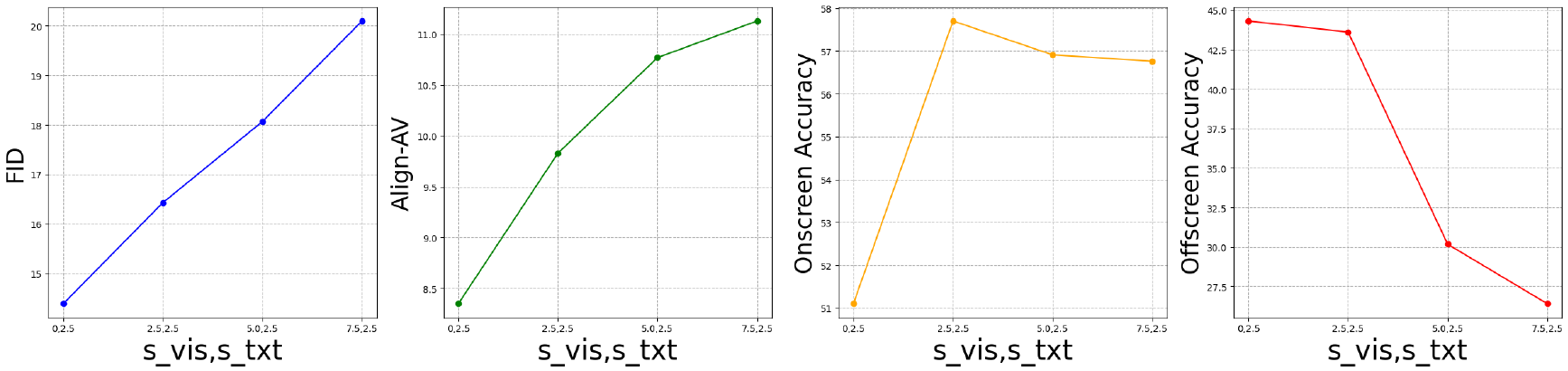}}
    \caption{$s_{vis} \in [0.0,2.5,5.0,7.5] $ and $s_{txt}$ = 2.5}
    \label{fig:vis_vary}
    \vspace{-3mm}
\end{figure}

We analyze the effect of fixing one weighting parameter during inference while uniformly varying the other. In Fig.~\ref{fig:vis_vary}, we present the metrics when $s_{\text{txt}}$ is fixed at 2.5, and $s_{\text{vis}}$ is incrementally increased from 0.0 to 7.5 in steps of 2.5. We observe a degradation in the FID, indicating a divergence from the real test distribution, alongside an improvement in audio-visual (AV) alignment. While on-screen accuracy reaches a plateau, offscreen accuracy decreases significantly, indicating in a stronger bias towards the visual modality.

\begin{figure}[h]
    \centerline{\includegraphics[width=0.45\textwidth]{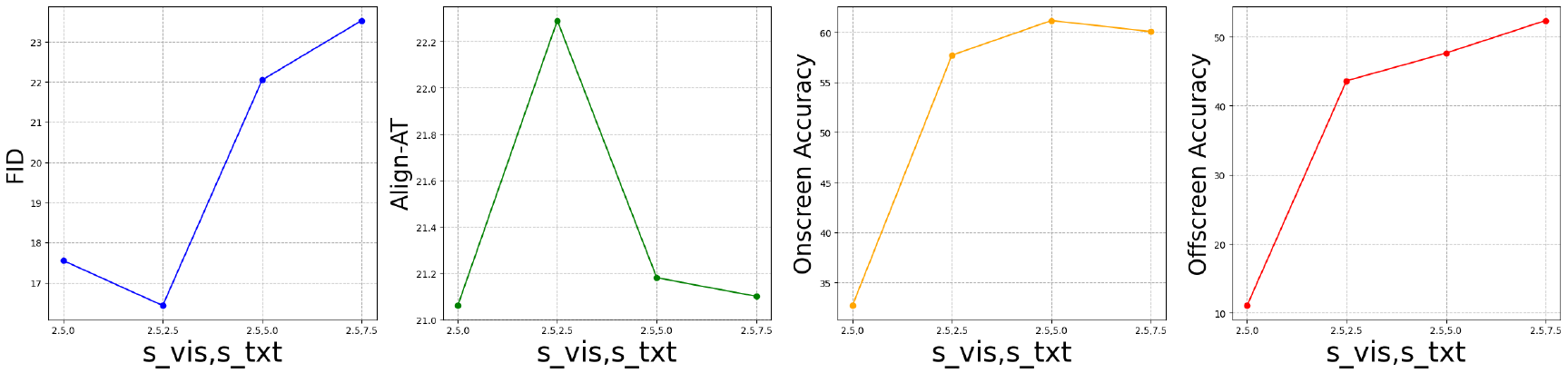}}
    \caption{$s_{vis} = 2.5 $ and $s_{txt} \in [0.0,2.5,5.0,7.5] $}
    \label{fig:txt_vary}
    \vspace{-3mm}
\end{figure}

Conversely, as shown in Fig.~\ref{fig:txt_vary}, when $s_{\text{vis}}$ is fixed at 2.5 and $s_{\text{txt}}$ is uniformly increased from 0.0 to 7.5 in steps of 2.5, the trends differ slightly. We observe a similar degradation in FID, but the audio-text (AT) alignment peaks at $s_{\text{txt}} = 2.5$. Additionally, both on-screen and offscreen accuracies consistently increase, enabling the model to effectively incorporate both on-screen and offscreen concepts.

\

\subsection{Analysing $w_l$.}

\begin{figure}[tb]
    \centerline{\includegraphics[width=0.45\textwidth]{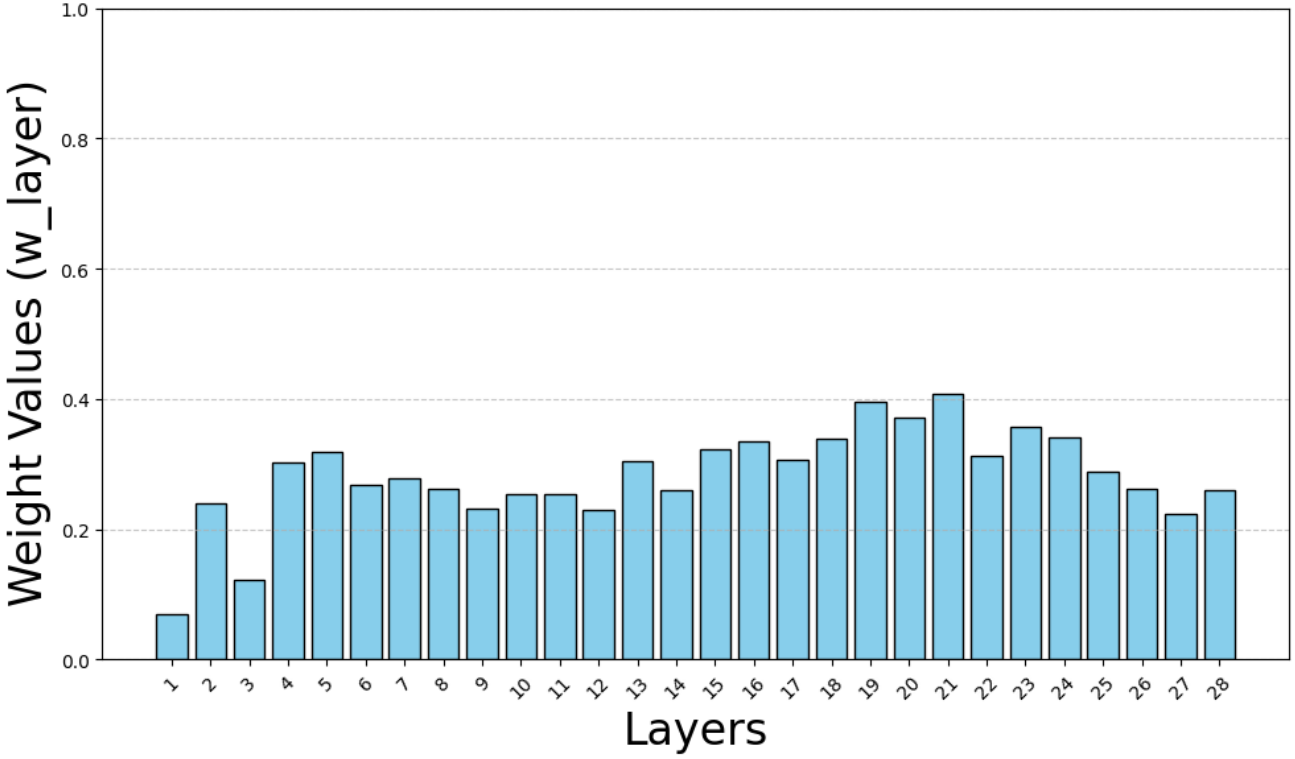}}
    \caption{Bar-plot for layer-wise cross-attention weights $w_l$}
    \label{fig:w_l}
    \vspace{-3mm}
\end{figure}

In Fig.~\ref{fig:w_l}, we plot the cross-attention weights for text and audio modality across each layer (or block) of our trained model. Lower text weights indicate a greater contribution from visual cross-attention (because visual weight is 1-$w_l$). 
This suggests that the model relies more on visual-semantic and temporal embeddings to effectively guide the fine-grained audio patch representations.

\begin{figure*}[t]
    \centering
    \includegraphics[width=0.99\textwidth]{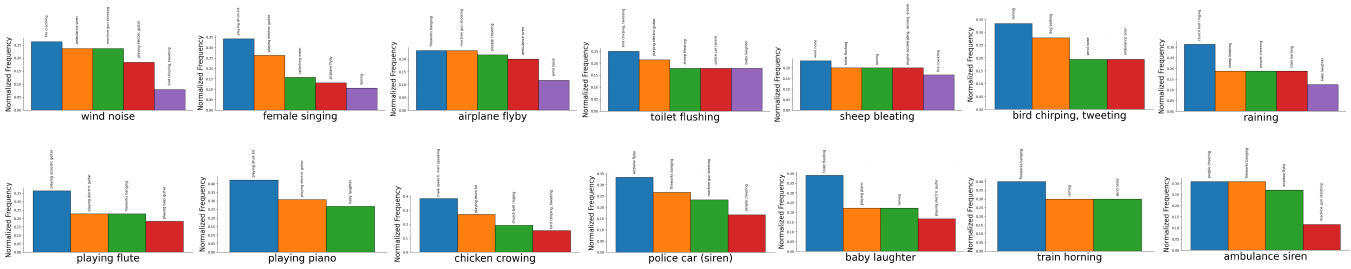}
    \caption{Distribution of onscreen and offscreen sounds. For each onscreen class, we plot the frequency distribution of off-screen sounds.}
    \vspace{-5mm}
    \label{fig:data_dist}
\end{figure*}

\section{Additional dataset details of VinTAGe-Bench}
\begin{figure}[tb]
    \centerline{\includegraphics[width=0.49\textwidth]{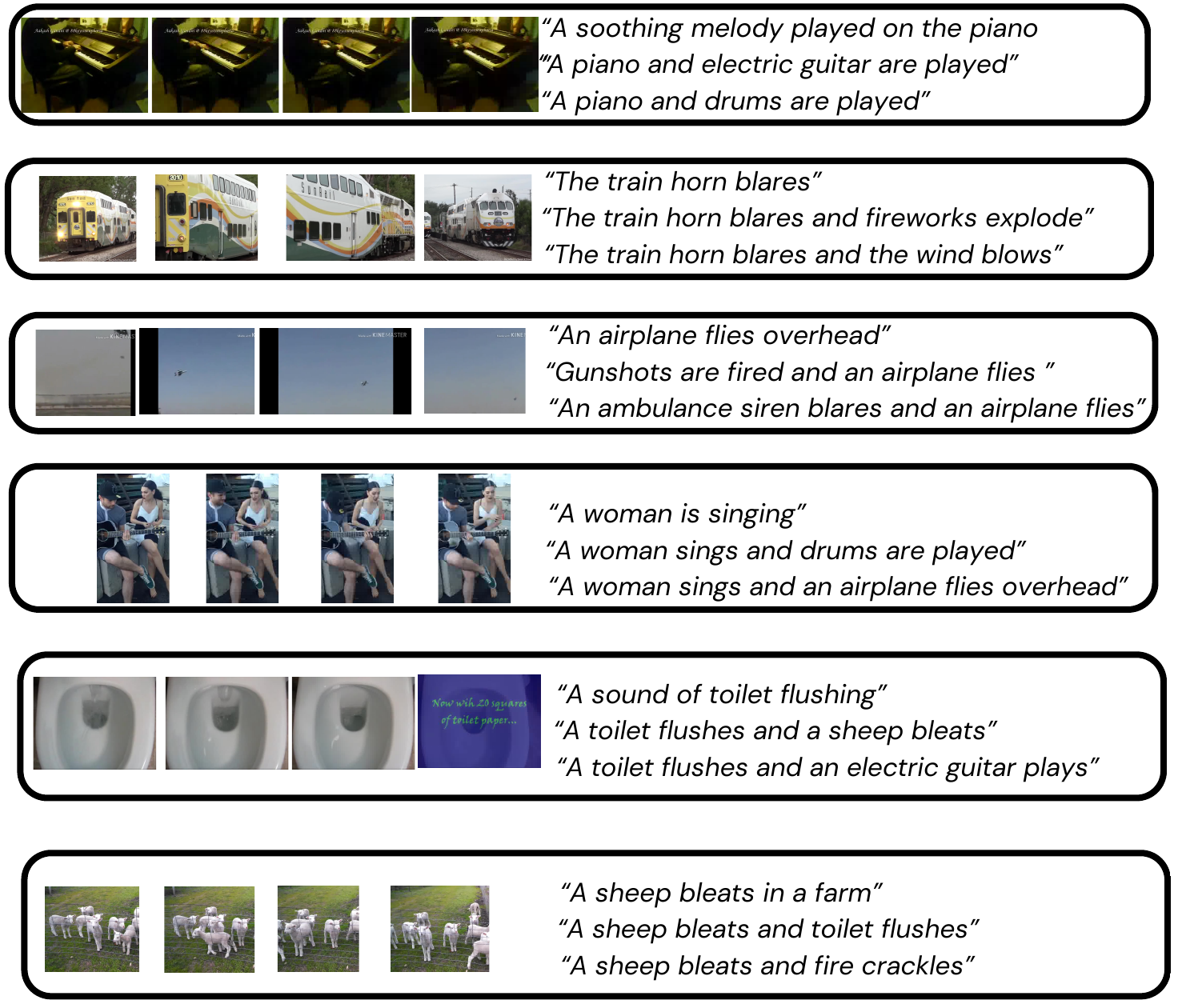}}
    \caption{More examples of VinTAGe-Bench}
    \label{fig:vintage_eg}
\end{figure}

\noindent \textbf{Dataset statistics of VinTAGe-Bench.} 
The newly curated dataset, VinTAGe-Bench, consists of 14 on-screen classes and 24 off-screen classes. In Fig.~\ref{fig:data_dist}, we present bar plots showing the frequency distribution of off-screen class categories for each on-screen sound. Each on-screen class includes at least three unique off-screen sounds. We provide additional examples of VinTAGe-Bench in Fig.~\ref{fig:vintage_eg}.

\noindent \textbf{Collection and annotations.} 
We selected videos from the VGGSound test subset. \\
\noindent \underline{\textit{Video Selection}.} First, we began with 20 categories from the VGGSound test set and removed videos where the class object was not visually present for the majority of the time, contained excessive noise, were animated, consisted of static images, or had sudden changes. Next, we removed classes with fewer than five videos, resulting in 14 classes.

\noindent \underline{\textit{Off-screen Classes}.} For each video category, we selected four off-screen classes (from the VGGSound categories) based on real-world co-occurrence and non-co-occurrence. For example, for the on-screen \textit{female singing} class, we selected co-occurring off-screen sounds such as \textit{playing drums} and \textit{playing electric guitar}, and non-co-occurring classes like \textit{airplane flying} and \textit{raining}. This approach enhances the robustness of our testing.

\noindent \underline{\textit{Annotation}.} For each video, we selected two off-screen categories and one no off-screen scenario from the offscreen categories for the video category. We then generated text captions by annotating three scenarios: (1) no off-screen sounds (e.g., \textit{"A soothing melody played on the piano"}), (2) with the first off-screen sound (\textit{"A piano and electric guitar are played"}), and (3) with the second off-screen sound (\textit{"A piano and drums are played"}).

\noindent \underline{\textit{Mixing Audios}.} For each data point with an off-screen sound, we selected a random audio from the VGGSound test set (of this off-screen sound category) and mixed it with the original video audio. To ensure a good representation of both audio samples, we considered the pressure levels of both audios and mixed them so that neither overpowers the other~\cite{ghosal2023tango,tokozume2017learning}. Additionally, we found that mixing audios with similar frequencies or when the on-screen sound is very busy made it difficult to recognize both sounds; therefore, we excluded these scenarios.

Finally, we obtained 636 data points consisting of 212 unique videos with 14 on-screen and 24 off-screen sound categories.

\section{Subjective test}
\begin{figure}[tb]
    \centerline{\includegraphics[width=0.49\textwidth]{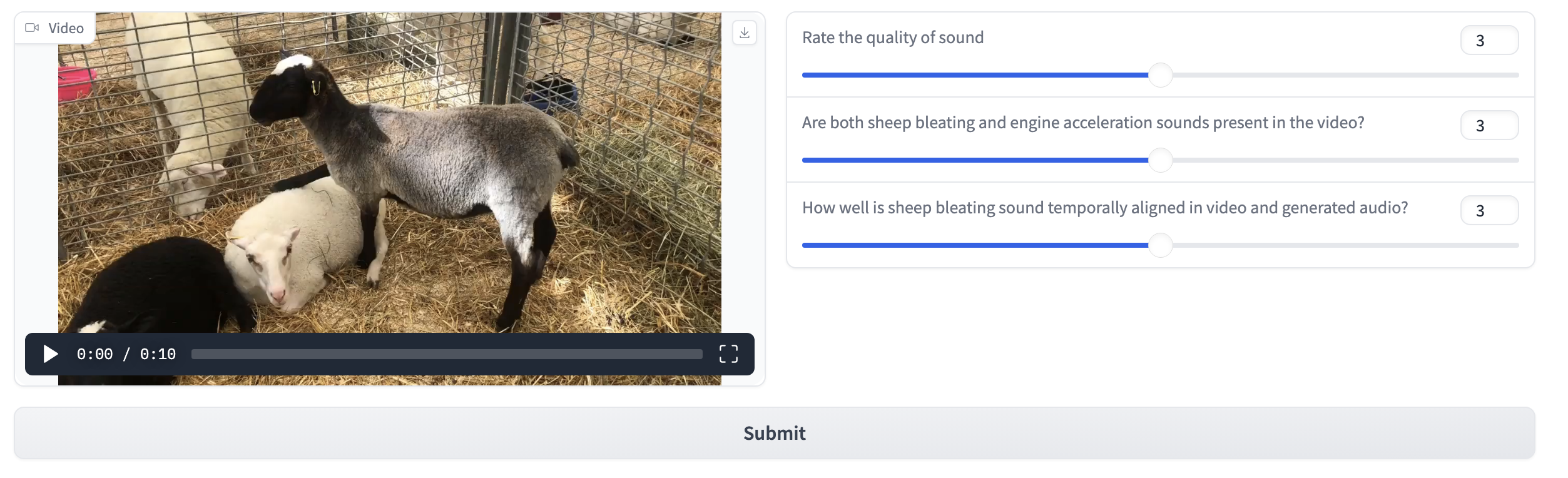}}
    \caption{Interface for subjective study.}
    \label{fig:subjective}
    \vspace{-3mm}
\end{figure}

We conducted user subjective studies on two datasets, VinTAGe-Bench and VGGSound. The interface for VinTAGe-Bench is shown in Fig.~\ref{fig:subjective}. We asked the participants to rate based on three criteria:

\begin{itemize}
    \item \textbf{Quality}: Rate the quality of the generated audio on a scale from 1 to 5.
    \item \textbf{Faithfulness}: Are both \textit{\{off-screen\}} and \textit{\{on-screen\}} sounds audible in the audio? Rate this faithfulness from 1 to 5.
    \item \textbf{On-screen Temporal Alignment}: How well does the \textit{\{on-screen\}} sound temporally align with the video? Rate this faithfulness from 1 to 5.
\end{itemize}

We calculated the mean for each question across all participants and reported the mean opinion scores in our results.
For the subjective test on VGGSound, we omitted question 2 and asked the same questions.

\section{Real-World Application}

\begin{figure}[tb]
    \centerline{\includegraphics[width=0.48\textwidth]{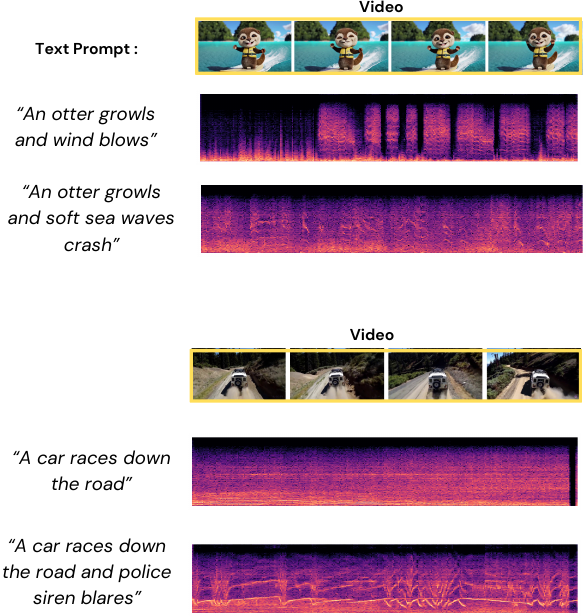}}
    \caption{VintAGe generated audio (spectrograms) for SORA videos. The input video and text are also shown.}
    \label{fig:sora}
\end{figure}

Our model is capable of generating audio for data outside the training distribution. For instance, we present examples in Fig.~\ref{fig:sora} of generated audio for OpenAI's SORA videos~\cite{videoworldsimulators2024}. Our model successfully generates realistic and holistic audio.

\section{Limitations and Future Work}

\noindent \textbf{Complex Scenarios}: Our model uses video energy (mean optical flow magnitudes) to temporally guide the generated audio, which works effectively in most scenarios. However, this approach may not suffice in complex situations involving multiple sound sources. Relying on mean optical flow magnitudes, the model might struggle to individually associate the temporal alignment of each sound source. In future work, we aim to explore object-wise temporal energy and incorporate spatial and temporal frame features to handle such complex scenarios more effectively.

\noindent \textbf{On-screen Temporal Alignment Metrics}: Existing audio-visual temporal alignment metrics~\cite{yariv2024diverse,luo2023difffoley} implicitly or explicitly assume that the audio has a single source or includes accompanying related sounds (e.g., frog croaking with rain falling sounds). In our case, however, we deal with mixed audio containing both on-screen and (maybe unrelated) off-screen sounds, rendering previous metrics unreliable for our purposes. Therefore, we rely on subjective tests to assess this metric. One possible solution is to extract the on-screen sounds using audio separation models~\cite{liu2023separate}, but we found that these models are not perfect and may miss or incorrectly separate the on-screen sounds. In future work, we plan to develop objective metrics to measure the temporal alignment of on-screen sounds with the visual input.

Similar to other generative models, we would also like to highlight the potential negative impacts of our method, such as generating synthetic audio that could be misused. It is crucial to implement safeguards to prevent the creation of harmful or sensitive audio content.

\end{document}